\documentclass[runningheads]{llncs}
\usepackage{graphicx}

\usepackage{tikz}
\usepackage{comment}
\usepackage{amsmath,amssymb} 
\usepackage{color}
\usepackage{multirow}
\usepackage{bbm}
\usepackage{booktabs}
\usepackage{wrapfig,lipsum}
\usepackage{adjustbox}

\usepackage{soul} 
\usepackage{pifont}
\newcommand{\cmark}{\ding{51}}%
\newcommand{\xmark}{\ding{55}}%
\usepackage[colorlinks]{hyperref}
\definecolor{Tdgreen}{rgb}{0,0.4,0.7}
\hypersetup{citecolor=Tdgreen}
\usepackage[symbol]{footmisc}

\usepackage[accsupp]{axessibility}  %

\definecolor{forestgreen}{rgb}{0.13, 0.55, 0.13}

\newcommand{\Dongqing}[1]{\textcolor{cyan}{\textbf{[Dongqing: #1]}}}

\begin{document}
\pagestyle{headings}
\mainmatter
\def\ECCVSubNumber{2149}  

\title{SPot-the-Difference Self-Supervised Pre-training for Anomaly Detection and Segmentation}

\titlerunning{SPot-the-Difference Representation Learning}
\author{Yang Zou\inst{1}, Jongheon Jeong\inst{2}\thanks{~~work done during an Amazon internship}, Latha Pemula\inst{1} \\Dongqing Zhang\inst{1}, Onkar Dabeer\inst{1}}
\authorrunning{Y. Zou et al.}
\institute{$^1$AWS AI Labs ~~~~~ $^2$KAIST \\
\email{\{yanzo,lppemula,zdongqin,onkardab\}@amazon.com, jongheonj@kaist.ac.kr}}
\maketitle

\begin{abstract}
 Visual anomaly detection is commonly used in industrial quality inspection. In this paper, we present a new dataset as well as a new self-supervised learning method for ImageNet pre-training to improve anomaly detection and segmentation in 1-class and 2-class 5/10/high-shot training setups. We release the Visual Anomaly (VisA) Dataset consisting of 10,821 high-resolution color images (9,621 normal and 1,200 anomalous samples) covering 12 objects in 3 domains, making it the largest industrial anomaly detection dataset to date. Both image and pixel-level labels are provided. We also propose a new self-supervised framework - SPot-the-difference (SPD) - which can regularize contrastive self-supervised pre-training, such as SimSiam, MoCo and SimCLR, to be more suitable for anomaly detection tasks. Our experiments on VisA and MVTec-AD dataset show that SPD consistently improves these contrastive pre-training baselines and even the supervised pre-training. For example, SPD improves Area Under the Precision-Recall curve (AU-PR) for anomaly segmentation by 5.9\% and 6.8\% over SimSiam and supervised pre-training respectively in the 2-class high-shot regime.
 We open-source the project at \url{http://github.com/amazon-research/spot-diff}.

\end{abstract}

\section{Introduction}
\begin{figure}[!t]
  \centering
  \includegraphics[width=0.95\linewidth]{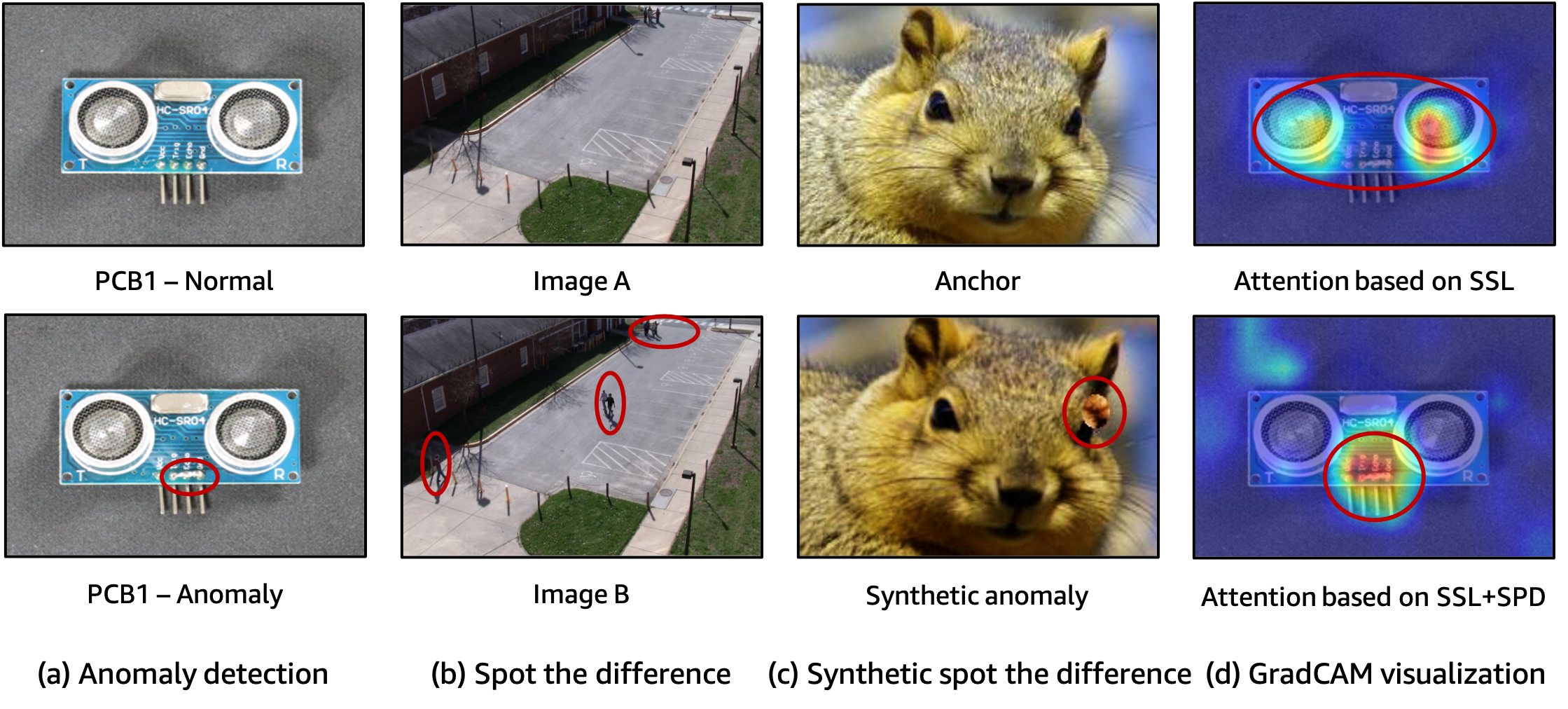}
  \caption{(a) Normal and anomalous samples of VisA - PCB1 with real defect (molten metal), anomaly highlighted by red ellipse; (b) A pair of images for the spot-the-difference (SPD) puzzle \cite{jhamtani2018learning}; (c) An anchor image and its variant augmented by SmoothBlend for synthetic spot-the-difference; (d) GradCAM attention visualization for PCB1 - Anomaly image based on self-supervised ImageNet pre-training w/wo proposed SPD. With SPD, attention is more focused on the local defects.}
  \label{fig:teaser}
\end{figure}

Visual surface anomaly detection and segmentation identify and localize defects in industrial manufacturing \cite{bergmann2021mvtec}. While anomaly detection and segmentation are instances of image classification and semantic segmentation problems, respectively, they have unique challenges. First, defects are rare, and it is hard to obtain a large number of anomalous images. Second, common types of anomalies, such as surface scratches and damages, are often small. Fig. ~\ref{fig:teaser} (a) gives an example. Third, manufacturing is a performance sensitive domain and usually requires highly accurate models. Fourth, inspection in manufacturing spans a wide range of domains and tasks, from detecting leakages in capsules to finding damaged millimeter-sized components on a complex circuit board. 

Upon the aforementioned challenges, previous surface anomaly detection models have been typically trained for a particular object and require re-training for different ones. For each object, there are only slight global differences in lighting and object pose/positions across images while the diversity in the defects on objects is large. 
Moreover, due to the rarity of anomalous data, there has been a predominant focus on 1-class anomaly detection, which only requires normal images for model training \cite{caron2020unsupervised,cohen2020sub,defard2021padim,li2021cutpaste,Roth_2022_CVPR,yi2020patch}. In mature manufacturing domains, anomalous samples are also available and sometimes sufficient. In such cases, one can improve over 1-class methods with a standard 2-class 
model \cite{cui2019class,feng2021few,gornitz2013toward,lin2017focal} by incorporating the anomalous data in training, which is in fact a well-established practice in 
commercial visual inspection AI services \cite{LfV,VIAI}.
For both setups, existing state-of-the-art methods for surface anomaly detection commonly leverage supervised representations pre-trained on ImageNet \cite{deng2009imagenet}, 
either as feature extractors \cite{defard2021padim,Roth_2022_CVPR} or as initialization for fine-tuning on the target dataset \cite{li2021cutpaste,yi2020patch}.

Meanwhile, recent advances in self-supervised learning (SSL) have shown that pre-trained representations learned without categorical labels might be a better choice for transfer learning compared to those from supervised in object detection and segmentation \cite{chen2020simple,chen2021exploring,he2020momentum}. However, their application to anomaly detection and segmentation is underdeveloped. SSL for surface anomaly detection was explored in CutPaste \cite{li2021cutpaste} to learn representation from downstream images for each specific object. However, such representations hardly generalize to different objects and can lead to overfitting in a practical setting where only 1-20 normal samples are available. Also, there are previous works focusing on SSL for high-level semantic anomaly detection such as cat among a distribution of dogs \cite{cook2020consult,davis2006relationship,saito2015precision}. However, as \cite{ruff2021unifying} pointed out, surface anomaly detection aims to spot the low-level textual anomalies such as scratch and crack which has challenges different from semantic anomaly detection. Until now, the universal self-supervised pre-trained representation with good generalization ability have not yet been attempted for surface anomaly detection and segmentation.

Regarding the evaluation protocol, the community has been experiencing the lack of challenging benchmarks. The popular MVTec Anomaly Detection (AD) benchmark \cite{bergmann2021mvtec} is  saturating with the Area Under the Receiver Operating Characteristic (AU-ROC) approaching $\sim$95\% \cite{defard2021padim,li2021cutpaste}, and the benchmark is limited to the 1-class setup. But the anomaly detection problems in practice is still far from solved, demanding new datasets and metrics that better represent the real-world. In this paper, we introduce a new challenging Visual Anomaly (VisA) dataset. VisA is collected to present several new characteristics: objects with complex structures such as printed circuit board (PCB), multiple instances with different locations in a single view, 12 different objects spanning 3 domains, and multiple anomaly classes (up to 9) for each object. VisA contains 10,821 high-resolution color images - 9,621 normal and 1,200 anomalous - with both image and pixel-level labels. To our best knowledge, VisA is currently the largest and most challenging public dataset for anomaly classification and segmentation. Moreover, to cover different use cases in practice, we establish benchmarks not only in standard 1-class training setup but also 2-class training setups with 5/10/high-shot. For evaluation, we propose to use Area Under the Precision-Recall curve (AU-PR) in combination with standard AU-ROC. In the imbalanced defect dataset, AU-ROC might present inflated view of performances and AU-PR is more informative to measure anomaly detection performance \cite{cook2020consult,davis2006relationship,saito2015precision}.

In addition to an improved dataset, we also explore self-supervision to improve anomaly detection. As we argue below, our hypothesis is that previous contrastive SSL methods \cite{chen2020simple,chen2021exploring,he2020momentum} are sub-optimal to transfer learning for anomaly detection. Specifically, SimCLR, MoCo and other methods regard globally augmented images of a given image as one class and other images in the same batch as negative classes. Transformations, such as cropping and color jittering, are applied globally to the anchor for positives generation. The InfoNCE or cosine similarity losses \cite{chen2020simple,chen2021exploring,he2020momentum} encourage invariance to these global deformations, and capturing semantic information instead of local details \cite{geirhos2020on}. However, anomaly detection relies on local textual details to spot defects. Thus the subtle and local intra-object (or intra-class) differences are important but not well modeled by previous methods. Figure \ref{fig:teaser} (d) illustrates the sub-optimality in one of the previous SSL methods using the GradCAM attention map \cite{selvaraju2017grad}. As far as we know, improving representations by self-supervision for better downstream anomaly detection/segmentation has not been studied before and we explore this angle. 

Inspired by the spot-the-difference puzzle shown in Fig. \ref{fig:teaser} (b), we propose a contrastive SPot-the-Difference (SPD) training to promote the local sensitivity of previous SSL methods. In the puzzle, players need to be sensitive to the subtle differences between the two globally alike images, which is similar to anomaly detection. In the contrastive SPD training, as shown in Fig. \ref{fig:teaser} (c), a novel augmentation called SmoothBlend is proposed to produce the local perturbations on SPD negatives for synthetic spot-the-difference. The (locally) augmented images are regarded as negatives, which is different from regarding (globally) augmented images as positives in SimCLR/MoCo. Moreover, weak global augmentations, such as weak cropping and color jittering, are also applied to the SPD negatives as anomaly detection should spot defects under slight global changes in lighting and object pose/position. Additionally, to prevent models from using the slight global changes as shortcuts to differentiate negatives, SPD positives are generated by applying weak global augmentations on the anchor. Lastly, SPD training minimizes the feature similarities between SPD negative pairs while maximizing the similarities between SPD positives, which encourages models to be locally sensitive to anomalous patterns and invariant to slight global variations.

Our main contributions are as follows:
\begin{enumerate}
    \item We propose a new VisA dataset, 2$\times$ larger than MVTec-AD, with both image and pixel-level annotations. It spans 12 objects across 3 domains, with challenging scenarios including complex structures in objects, multiple instances and object pose/location variations. Moreover, we establish both 1-class and 5/10/high-shot 2-class benchmarks to cover different use cases.
    \item To promote the local sensitivity to anomalous patterns, a SPot-the-Difference (SPD) training is proposed to regularize self-supervised ImageNet pre-training, which benefits their transfer-learning ability for anomaly detection and localization. As far as we know, we are the first one to explore self-supervised pre-training on large-scale datasets for surface defect detection tasks. 
    \item Compared to strong self-supervised pre-training baselines such as SimSiam, MoCo and SimCLR, extensive experiments show our proposed SPD learning improves them for better anomaly detection and segmentation. We also show the SPD improves over supervised ImageNet pre-training for both tasks. 
\end{enumerate}
\section{Related Works}
\noindent\textbf{Unsupervised Anomaly Detection and Segmentation} use only normal samples to train models, which have drawn extensive attention. Many recent methods are proposed to detect low-level texture anomalies \cite{ruff2021unifying}, such as scratches and cracks, which are common cases in industrial visual inspection \cite{Deng_2022_CVPR,Ristea-CVPR-2022,RudWeh2022,yi2020patch}. SPADE \cite{defard2021padim} and PatchCore \cite{Roth_2022_CVPR} extract features at patch level and use nearest neighbor methods to classify patches and images as anomalies. PaDiM \cite{defard2021padim} learns a parametric distribution over patches for anomaly detection. CutPaste \cite{li2021cutpaste} learns a representation based on images augmented by cut-and-pasted patches. The supervised ImageNet models are used in these methods either as feature extractors or initialization for fine-tuning. However, self-supervised pre-training on large-scale datasets is an unexplored area for quality inspection applications. In addition, several works \cite{reiss2021mean,sohn2021learning,tack2020csi,pmlr-v80-ruff18a} focus on high-level semantic anomaly detection. As mentioned in \cite{ruff2021unifying}, semantic anomaly detection approaches can be less effective for texture anomaly detection as their challenges are different.

\noindent\textbf{Self-Supervised Learning (SSL)} have gathered momentum in the last 5 years. Several surrogate tasks have been proposed for self-supervision, such as image colorization \cite{zhang2016colorful}, rotation prediction \cite{gidaris2018unsupervised}, jigsaw puzzles \cite{noroozi2016unsupervised}. Recently, multi-view based methods such as MoCo \cite{he2020momentum}, SimCLR \cite{chen2020simple}, SimSiam \cite{chen2021exploring} and BYOL \cite{grill2020bootstrap} present better or comparable performances than supervised pre-training in transfer learning tasks including image classification, object detection \cite{Yang_2021_CVPR} and semantic segmentation \cite{wang2021dense}. Moreover, to promote spatial details of representations for localization tasks, several approaches proposed to encourage the invariance of patch features to global augmentations \cite{wang2021dense,xie2021detco,liu2020self,chen2021multisiam}, although they may not lead to local sensitivity to tiny defects. As far as we know, none of these works explored their generalization ability to surface defect detection tasks. 

\begin{figure}[!t]
 \centering
\includegraphics[width=.8\linewidth]{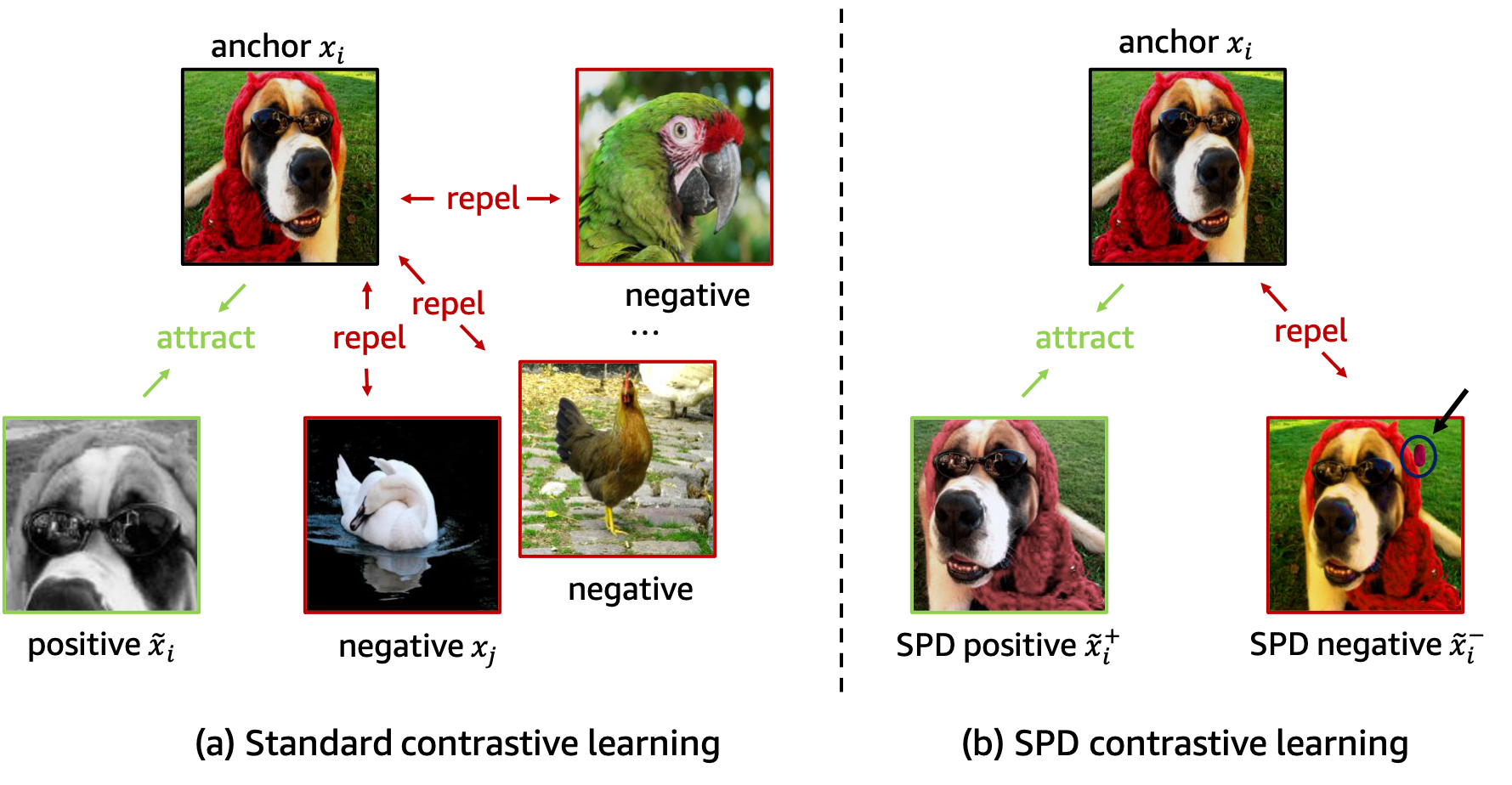}
\caption{(a) Contrastive learning in SimCLR, MoCo and SimSiam; (b) Contrastive learning in SPD training. Local deformation in SPD negative is highlighted by circle.}
\label{fig:ssl_comp}
\end{figure}

\section{SPot-the-Difference (SPD) Regularization}
To promote local sensitivity of standard self-supervised contrastive learning, we propose a contrastive SPot-the-Difference (SPD) regularization. As mentioned earlier, SPD aims to increase model invariance to slight global changes by maximizing the feature similarity between an image and its weak global augmentation, while forcing dissimilarity for local perturbations, as shown in Fig. \ref{fig:ssl_comp} (b). In the following, we first present background in contrastive learning, and then the augmentations used in SPD followed by the learning with SPD.
\subsection{Background on Self-supervised Contrastive Learning}
\begin{figure}[!t]
\centering
\includegraphics[width=\linewidth]{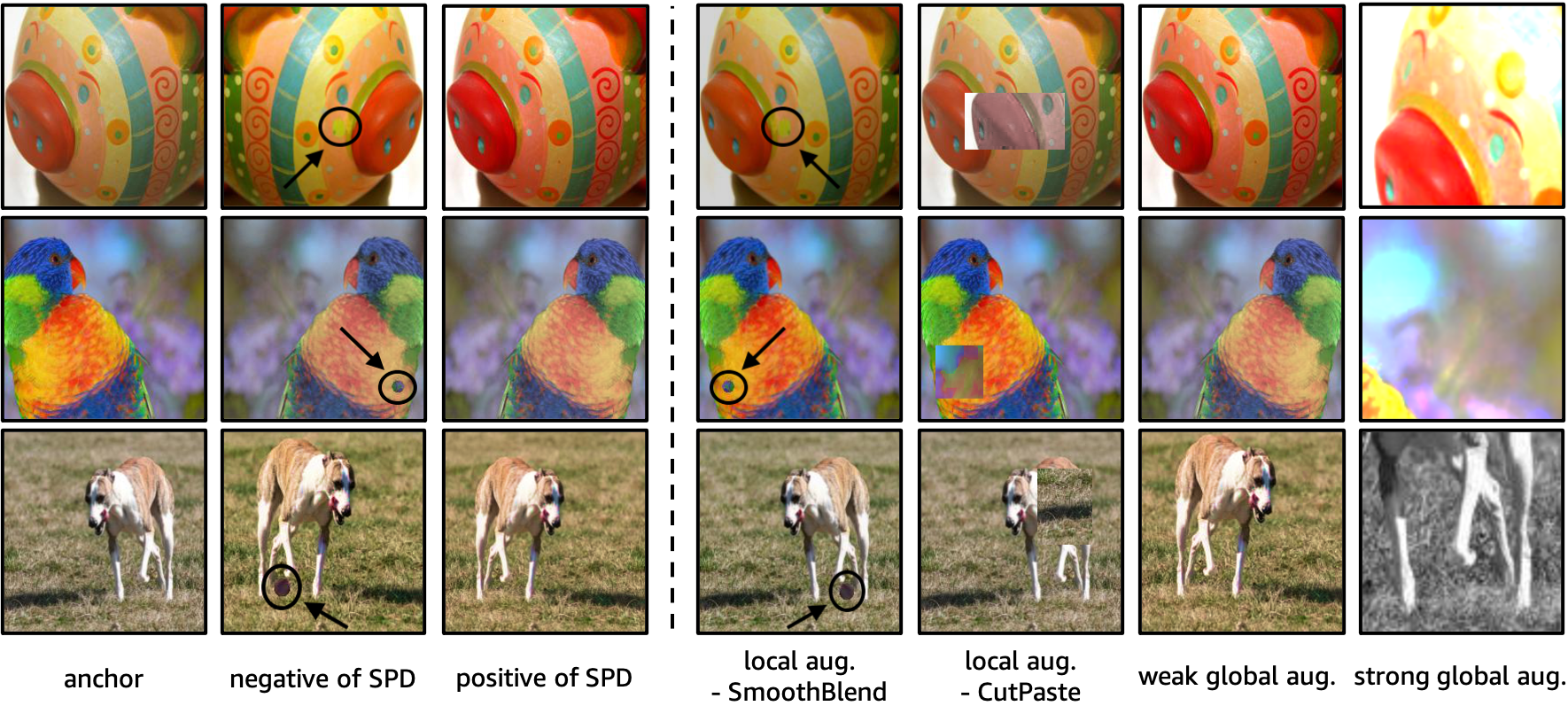}
\caption{(a) Samples for synthetic spot-the-difference; (b) Augmentation comparison}
\label{fig:aug_compare}
\end{figure}

Many self-supervised learning methods, such as SimCLR \cite{chen2020simple} and MoCo \cite{he2020momentum}, are based on contrastive learning. As shown in Fig. \ref{fig:ssl_comp} (a), given an image, these methods maximize the feature similarity between two strongly augmented samples $x_i$ and $\hat{x}_i$ while minimizing the similarities between the anchor $x_i$ and other images $x_j$'s in the same batch of size $N$. Strong global augmentations, such as grayscaling, large cropping and strong color jittering, are used to get positives. Typically, an encoder extracts features $h_i,\hat{h}_i$ and $h_j$'s which are inputs to a multilayer perceptron (MLP) head. The MLP head extracts the L2 normalized embeddings $z_i,\hat{z}_i$ and $z_j$'s to compute the InfoNCE loss defined as follows.
\begin{equation}
    \mathcal{L}_{\mathrm{NCE}}(x_i,\hat{x}_i)= -\log\frac{\exp{(z_i\cdot \hat{z}_i/\tau)}}{\exp{(z_i\cdot \hat{z}_i/\tau)}+\sum_{j=1}^{N}\mathbbm{1}_{j\neq i}\exp{(z_i,z_j}/\tau)}
    \label{eqn:loss_nce}
\end{equation}
$\tau$ is a temperature scaling hyperparameter. In addition, SimSiam \cite{chen2021exploring} shows that self-supervised models can be trained even without negatives where only similarity modeling is implemented for positives.

\noindent\textbf{Remark:} Images augmented by most strong global transformations in SSL, such as grayscaling and large cropping, share semantics with anchor but with different local details (a dog v.s. a dog head). Thus to maximize their similarity, the features are forced to be invariant about local details and capture the global semantics. This is even enforced by minimizing similarities between anchor and different images in a batch as they have different global structures \cite{chen2020simple,ericsson2021self}. This further motivates us to promote local sensitivity in SSL for anomaly detection.
\subsection{Augmentations for SPD}
\noindent\textbf{Local augmentation:} In SPD, the locally deformed images, rather than other images of a batch in standard contrastive training, are used as negatives. SmoothBlend is proposed to produce local deformations. The first column in Fig. \ref{fig:aug_compare} (b) presents the samples augmented by SmoothBlend. It is implemented by a smoothed alpha blending between an image and a small randomly cut patch of the same image. Specifically, color jittering is applied to a cut patch. Then an all-zero foreground layer $u$ is created with the patch pasted to a random location. An alpha mask $\alpha$ is created where the pixels corresponding to the pasted patch are set to $1$ otherwise $0$, followed by a Gaussian blur. Finally, the augmented sample is obtained by $\bar{x} = (1-\alpha) \odot x + \alpha \odot u$. $\odot$ is the element-wise product. 

\noindent\textbf{Global augmentation:} To generate global variations for both SPD positives and negatives, we use weak global augmentation. Adding global variations to SPD is motivated by the potentially small global variations in realistic manufacturing environment, such as lighting, object positions, etc. To simulate such slight changes, we choose weak random cropping, Gaussian blurring, horizontal flipping and color jittering. Such weak global augmentations are different from strong transformations used in SimSiam, SimCLR and MoCo which is illustrated by last two columns in Fig. \ref{fig:aug_compare} (b). As we can see, there might be just $20\%$ overlap between the anchor and strongly augmented positive. If the network is designed to maximize the distance between negatives with only subtle changes while minimizing the distance between positives with largely global transformations, it is a confusing task which might harm representation learning for anomaly detection.

\noindent\textbf{Remark:} SmoothBlend is a smoothed version of CutPaste augmentation proposed in \cite{li2021cutpaste}. Both of them can be used to generate structural local deformations, illustrated by the first two columns in Fig. \ref{fig:aug_compare} (b). Unlike the sharp edges of the CutPaste patches, the local and subtle perturbations with smooth edges from SmoothBlend provides a challenging puzzle for models. 
\begin{figure}[!t]
 \centering
\includegraphics[width=\linewidth]{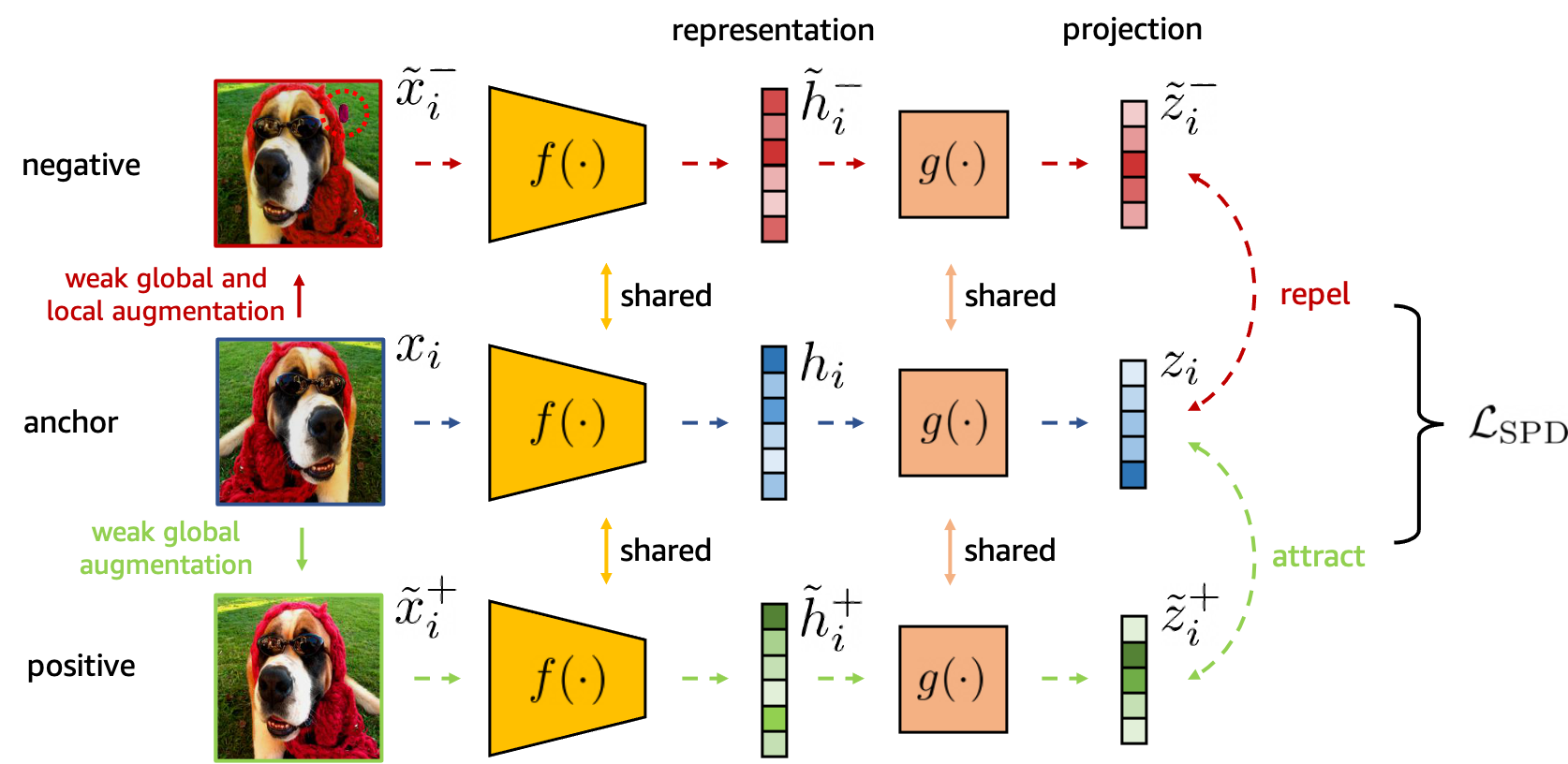}
\caption{The contrastive spot-the-difference learning}
\label{fig:spd_reg}
\end{figure}
\subsection{Training with SPD}
Based on the above augmentations, we propose the SPD learning illustrated by Fig. \ref{fig:spd_reg} with Fig. \ref{fig:aug_compare} (a) presents more SPD training samples. For an anchor image $x_i$, a negative $\tilde{x}_i^-$ is generated by applying weak global augmentations followed by SmoothBlend. The positive $\tilde{x}_i^+$  is produced by weak global transformations only. Then a shared feature extractor $f(\cdot)$ extracts the representations $h_i, \tilde{h}_i^-, \tilde{h}_i^+$ ($h_i$'s are used for downstream anomaly detection tasks). They are further inputted into a shared multilayer perceptron (MLP) $g(\cdot)$ to get the projections $z_i, \tilde{z}_i^-,\tilde{z}_i^+$. The cosine similarity between $z_i, \tilde{z}_i^-$ is minimized while similarity between $z_i, \tilde{z}_i^+$ is maximized. In summary, the SPD learning minimizes the following SPD loss. 
\begin{equation}
    \mathcal{L}_{\mathrm{SPD}}(x_i,\tilde{x}_i^-,\tilde{x}_i^+)=
\cos(z_i, \tilde{z}_i^-)-\cos(z_i, \tilde{z}_i^+).
    \label{eqn:l_spd}
\end{equation}

\noindent\textbf{Standard contrastive SSL with SPD:} Regularizing SSL with SPD is simple. Taking SimCLR as an example baseline, for a given image, SimCLR generates the anchor $x_i$ and positive $\hat{x}_i$ via strong global augmentations with other images $x_j$'s in the same batch as negatives. Then SPD positives $\hat{x}_i^+$ and negatives $\hat{x}_i^-$ are generated by SmoothBlend and weak global augmentations. The shared encoder and MLP head in SimCLR are used to extract the image feature projections for loss computation. Finally the network is trained by the following combined loss.
\begin{equation}
    \mathcal{L}(x_i,\hat{x}_i,\tilde{x}_i^-,\tilde{x}_i^+) = \mathcal{L}_{\mathrm{NCE}}(x_i,\hat{x}_i) + \eta \cdot \mathcal{L}_{\mathrm{SPD}}(x_i,\tilde{x}_i^-,\tilde{x}_i^+)
\end{equation}
Similary, we can apply SPD to MoCo. For SimSiam, $\mathcal{L}_{\mathrm{NCE}}(x_i,\hat{x}_i)$ loss is replaced by a cosine distance loss for positive pairs without considering negatives \cite{chen2021exploring}.

\noindent\textbf{Standard supervised pre-training with SPD:} With the class labels, standard supervised pre-trained features also capture global semantics to distinguish categories with less attention to local details, similar to SSL. Thus SPD could improve its local sensitivity. Specifically, on top of the last feature layer of the standard supervised model (ResNet-50 \cite{he2016deep}), an auxiliary classifier is added to classify if an augmented SPD image has a local perturbation or not, which is trained by cross-entropy loss. The backbone is shared to extract features.
\section{Visual Anomaly (VisA) Dataset}
\subsection{Dataset Description}
\begin{figure}[!t]
 \centering
\includegraphics[width=\linewidth]{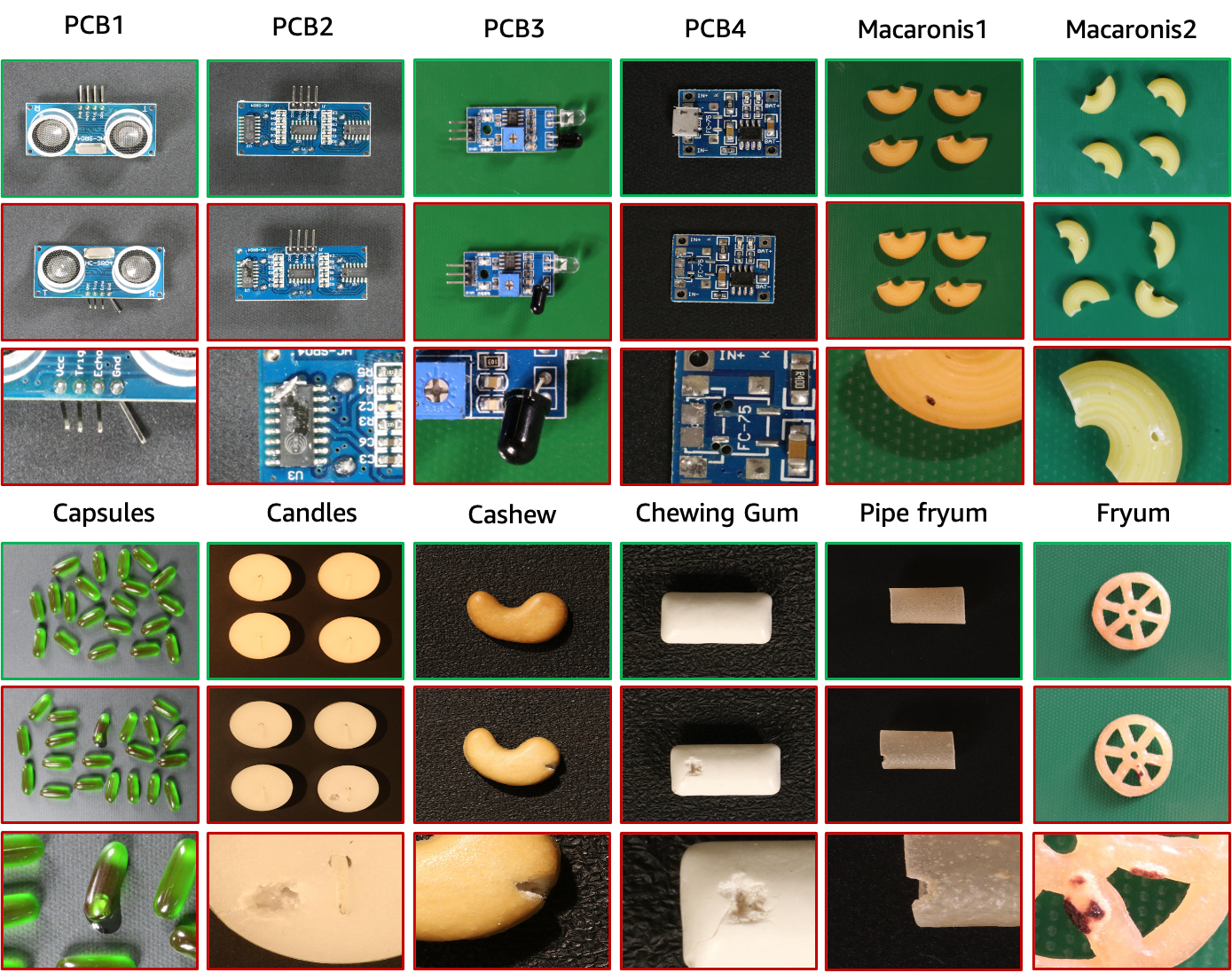}
\caption{Samples of VisA datasets. First row: normal images; Second row: anomalous images; Third row: anomalies viewed by zooming in.}
\label{fig:ava_samples}
\end{figure}

\begin{wraptable}[11]{r}{0.55\textwidth}
\setlength\abovecaptionskip{-1.8\baselineskip}
\scriptsize
\centering
\caption{Overview of VisA dataset}
\medskip
\label{tab:ava_stat}
\begin{adjustbox}{width=\linewidth}
\begin{tabular}{c|c|c|c|c}
\hline
                                                                               & Object   & \# normal & \# anomaly & \# anomaly \\
 &             & samples & samples & classes \\ \hline
\multirow{4}{*}{\begin{tabular}[c]{@{}c@{}}Complex \\ structure\end{tabular}}  & PCB1     & 1,004     & 100        & 4          \\
 & PCB2        & 1,001   & 100     & 4       \\
 & PCB3        & 1,006   & 100     & 4       \\
 & PCB4        & 1,005   & 100     & 7       \\ \hline
\multirow{4}{*}{\begin{tabular}[c]{@{}c@{}}Multiple \\ instances\end{tabular}} & Capsules & 602       & 100        & 5          \\
 & Candle      & 1,000   & 100     & 8       \\
 & Macaroni1   & 1,000   & 100     & 7       \\
 & Macaroni2   & 1,000   & 100     & 7       \\ \hline
\multirow{4}{*}{\begin{tabular}[c]{@{}c@{}}Single\\  instance\end{tabular}}   & Cashew      & 500     & 100     & 9       \\
 & Chewing gum & 503     & 100     & 6       \\ 
 & Fryum    & 500       & 100        & 8          \\
 & Pipe fryum  & 500     & 100     & 6       \\ \hline
\end{tabular}
\end{adjustbox}
\end{wraptable}
The VisA dataset contains 12 subsets corresponding to 12 different objects. Fig. \ref{fig:ava_samples} gives images in VisA. There are 10,821 images with 9,621 normal and 1,200 anomalous samples. Four subsets are different types of printed circuit boards (PCB) with relatively complex structures containing transistors, capacitors, chips, etc. For the case of multiple instances in a view, we collect four subsets: Capsules, Candles, Macaroni1 and Macaroni2. Instances in Capsules and Macaroni2 largely differ in locations and poses. Moreover, we collect four subsets including Cashew, Chewing gum, Fryum and Pipe fryum, where objects are roughly aligned. The anomalous images contain various flaws, including surface defects such as scratches, dents, color spots or crack, and structural defects like misplacement or missing parts. There are 5-20 images per defect type and an image may contain multiple defects. The defects were manually generated to produce realistic anomalies. All images were acquired using a $4,000\times6,000$ high-resolution RGB sensor. Both image and pixel-level annotations are provided. Table \ref{tab:ava_stat} gives the statistics of VisA dataset. 

\begin{figure}[!t]
 \centering
\includegraphics[width=.95\linewidth]{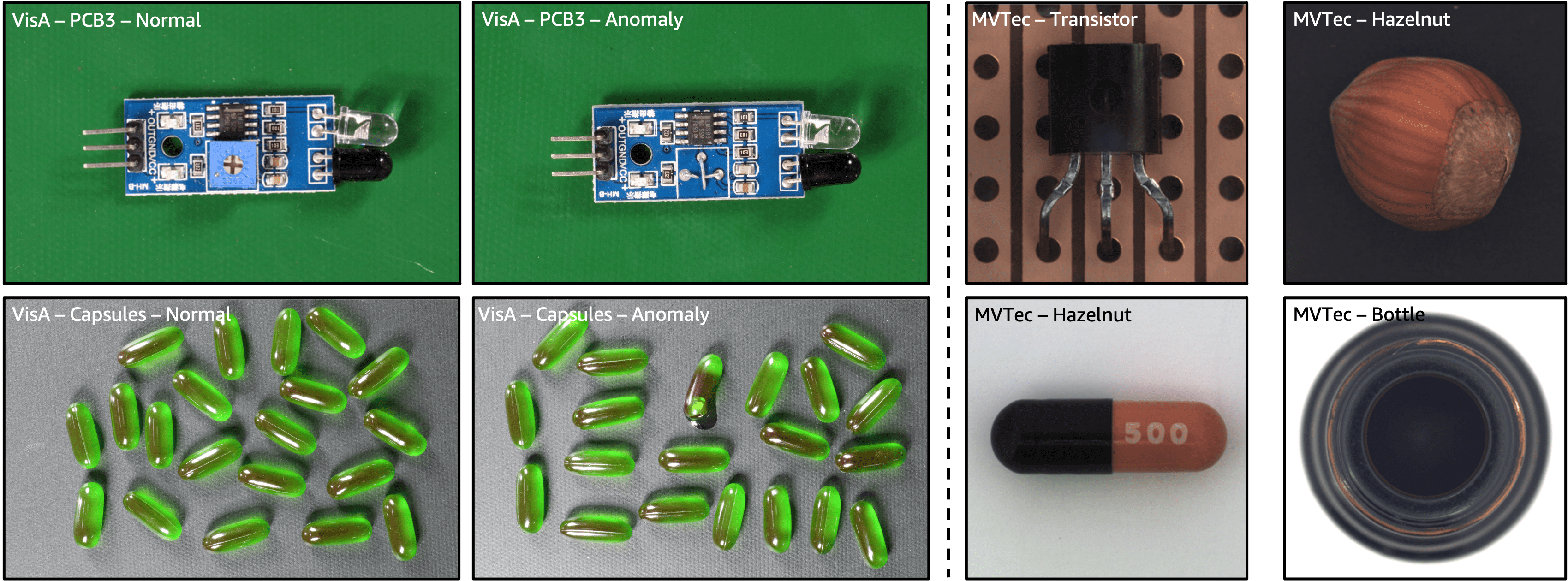}
\caption{Comparing VisA and MVTec-AD. VisA is more challenging due to the complex object structures, multiple instances, large variations of objects and scale.}
\label{fig:ava_vs_mvtec}
\end{figure}

Fig. \ref{fig:ava_vs_mvtec} illustrates the differences between VisA and MVTec-AD. First, VisA considers more complex structures, comparing the VisA - PCB3 with multiple electronic components to a single one of MVTec - transistor as an example. Second, multiple objects can appear in VisA (Capsules) as opposed to a single object in MVTec-AD. Third, large variation in object locations is covered by VisA (Capsules) while almost all objects in MVTec-AD are roughly aligned. Lastly, MVTec-AD has $5,354$ images and VisA is 2$\times$ larger with $10,821$ images.
\subsection{Evaluation Protocol and Metrics}
We establish three evaluation protocols for each of 12 objects in VisA dataset. First, following MVTec-AD 1-class protocol, we establish VisA 1-class protocol by assigning $90\%$ normal images to train set while $10\%$ normal images and all anomalous samples are grouped as test set. Second, we establish 2-class high/low-shot evaluation protocols as proxies for realistic 2-class setups in commercial products \cite{LfV,VIAI}. In high-shot setup, for each object, $60\%$/$40\%$ normal and anomalous images are assigned to train/test set respectively. For low-shot benchmark, firstly, $20\%$/$80\%$ normal and anomalous images are grouped to train/test set respectively. Then the k-shot (k=5,10) setup randomly samples k images from both classes in train set for training. The averaged performances over 5 random runs will be reported. Note that for both 1-class and 2-class training setups, test sets have samples from both classes. In addition, we report model performances averaged over all subsets of VisA and MVTec-AD in Sec. \ref{sec:exp}. The model performances for each subset are reported in Sec. \textcolor{red}{D} of supplementary.

For metrics, we report Area Under Precision-Recall curve (AU-PR) in combination with the Area Under Receiver Operator Characteristic curve (AU-ROC). AU-ROC is the most widely used metric for anomaly detection tasks \cite{defard2021padim,Roth_2022_CVPR,yi2020patch}. But as pointed out in \cite{cook2020consult,davis2006relationship,saito2015precision}, in imbalanced dataset where performance of minor class is more important, AU-ROC might provide an inflated view of performance which may cause challenges in measuring models' true capabilities. This is true for anomaly detection where anomalies are often rare. In \cite{bergmann2021mvtec}, the best method is Student-Teacher \cite{bergmann2020uninformed} with $92.2\%$ AU-ROC which seems to be close to perfection. However, it only gets $59.9\%$ AU-PR which is far-from satisfactory. The imbalance issue is more extreme in anomaly segmentation where normal pixels (negatives) can be tens/hundreds times more than anomalous pixels (positives). Even for a bad model, the false positive rate can be small due to numerous negatives, leading to a high AU-ROC. Thus we argue AU-PR is a better performance measurement. Our experiments also demonstrate this point.

\section{Experiments}\label{sec:exp}

\noindent\textbf{Datasets:} For self-supervised as well as supervised pre-training, we use ImageNet 2012 classification dataset \cite{deng2009imagenet}. ImageNet consists $1,000$ classes with $1.28$ million training images. For downstream tasks, in addition to our VisA dataset, we use MVTec-AD dataset \cite{bergmann2019mvtec} as a 1-class training benchmark. MVTec-AD contains $15$ sub-datasets with a total of $5,354$ images.

\noindent\textbf{Anomaly detection and segmentation algorithms:} To evaluate the transfer learning performances of different pre-training, we adopt the following algorithms for anomaly detection and segmentation. 

\noindent\emph{1-class anomaly classification/segmentation:} We leverage PaDiM \cite{defard2021padim} which is one of the top performing 1-class anomaly detection/localization methods. 

\noindent\emph{2-class anomaly classification/segmentation:} We train a standard binary ResNet \cite{he2016deep} as the supervised model for classification. A U-Net \cite{ronneberger2015u} is used as segmentation model. The focal loss \cite{lin2017focal} is used to overcome the data imbalance. 

\noindent\textbf{Implementation details:} Unless otherwise noted, we choose ResNet-50 as the major backbone. We adopt exactly the same hyperparameters in SimSiam, MoCo, SimCLR and supervised learning for pre-training. More implementation details are in the supplementary.

\subsection{SPD in high-shot 1-class/2-class Regimes}
\begin{figure}[!t]
\centering
\includegraphics[width=0.45\textwidth]{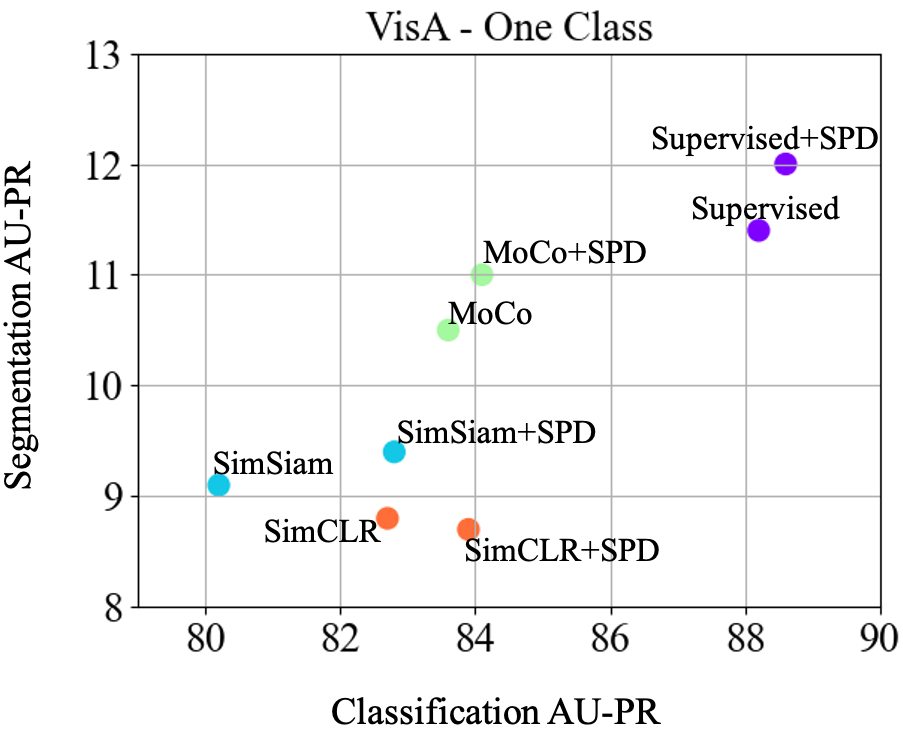}
\includegraphics[width=0.45\textwidth]{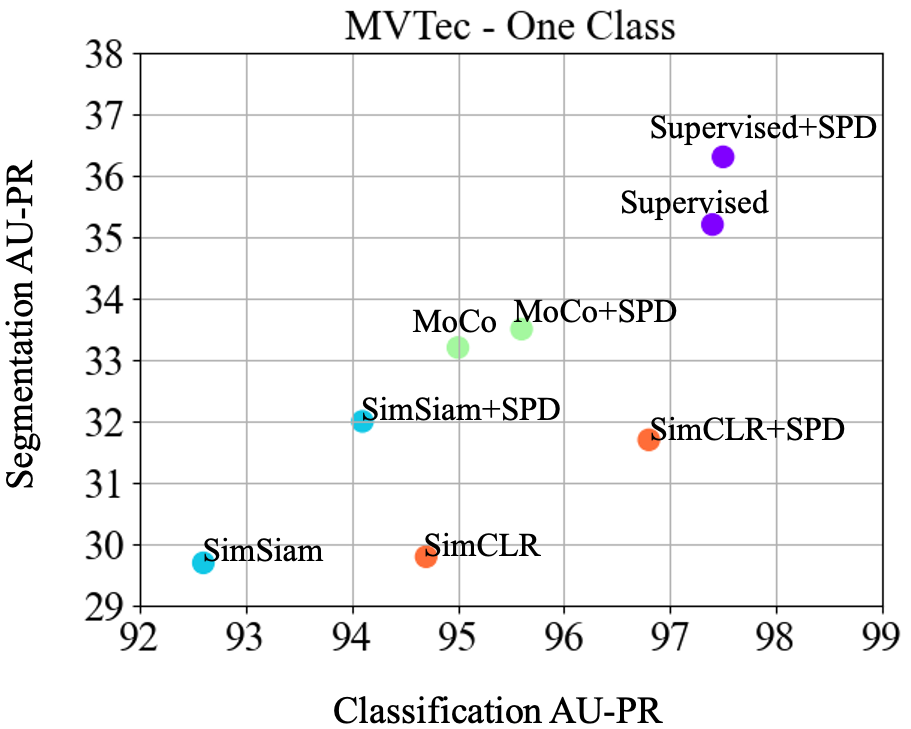}\\
 \caption{Scatter plots for various ImageNet pre-training models in 1-class setup.}
 \label{fig:scatter}
\end{figure}

\begin{table*}[!t]
	\centering
    \caption{1-class performance evaluation of various ImageNet pre-training options on VisA and MVTec-AD with PaDiM. Bold numbers refers to the highest score. In the brackets are the gaps to the ImageNet supervised/self-supervised pre-training counterpart. In green are the gaps of at least \textcolor{forestgreen}{+0.5} point.} 
    \label{tab:1cls}
	\setlength{\tabcolsep}{7pt}
	\resizebox{\linewidth}{!}{
	\centering
    \begin{tabular}{c|c|llll||llll}
    \hline
    \multicolumn{1}{l|}{\multirow{2}{*}{}} &
      \multirow{2}{*}{\begin{tabular}[c]{@{}c@{}}ImageNet\\ labels\end{tabular}} &
      \multicolumn{4}{c||}{VisA (1-class)} &
      \multicolumn{4}{c}{MVTec-AD (1-class)} \\ \cline{3-10} 
    \multicolumn{1}{l|}{} &
       &
      \multicolumn{2}{c|}{Classification} &
      \multicolumn{2}{c||}{Segmentation} &
      \multicolumn{2}{c|}{Classification} &
      \multicolumn{2}{c}{Segmentation} \\ \hline
                         &  & AU-PR & \multicolumn{1}{l|}{AU-ROC} & AU-PR & AU-ROC & AU-PR & \multicolumn{1}{l|}{AU-ROC} & AU-PR & AU-ROC \\ \hline
    Sup. pre-train       & \cmark & 88.2 & \multicolumn{1}{l|}{87.8}   & 11.4 & 93.1   & 97.4  & \multicolumn{1}{l|}{94.5}  & 35.2 & 94.4  \\ \hline\hline
    SimSiam              & \xmark & 80.2 & \multicolumn{1}{l|}{78.1}   & 9.1  & 93.1  & 92.6 & \multicolumn{1}{l|}{83.9}  & 29.7 & 92.1  \\
    +SPD                 & \xmark & 82.8 \textcolor{forestgreen}{(+2.6)} & \multicolumn{1}{c|}{81.2 \textcolor{forestgreen}{(+3.1)}}  & 9.4 (+0.3) & 92.7 (-0.4) & 94.1 \textcolor{forestgreen}{(+1.5)} & \multicolumn{1}{l|}{88.0 \textcolor{forestgreen}{(+4.1)}}  & 32.0 \textcolor{forestgreen}{(+2.3)} & 92.2 (+0.1) \\ \hline
    MoCo                 & \xmark & 83.6 & \multicolumn{1}{l|}{83.4}  & 10.5 & 93.4  & 95.0 & \multicolumn{1}{l|}{90.4}  & 33.2 & 93.4  \\
    +SPD                 & \xmark & 84.1 \textcolor{forestgreen}{(+0.5)} & \multicolumn{1}{l|}{83.0 (-0.4)}  & 11.0 \textcolor{forestgreen}{(+0.5)} & 93.5 (+0.1) & 95.6 \textcolor{forestgreen}{(+0.6)} & \multicolumn{1}{l|}{90.5 (+0.1)}  & 33.5 (+0.3) & 93.5 (+0.1) \\ \hline
    SimCLR               & \xmark & 82.7 & \multicolumn{1}{l|}{81.6}  & 8.8  & 89.7   & 94.7 & \multicolumn{1}{l|}{90.7}  & 29.8 & 92.1  \\
    +SPD                 & \xmark & 83.9 \textcolor{forestgreen}{(+0.8)} & \multicolumn{1}{l|}{82.6 \textcolor{forestgreen}{(+1.0)}}  & 8.7 (-0.1) & 89.9 (+0.2) & 96.8 \textcolor{forestgreen}{(+2.1)} & \multicolumn{1}{l|}{93.8 \textcolor{forestgreen}{(+3.1)}}  & 31.7 \textcolor{forestgreen}{(+1.9)} & 92.9 \textcolor{forestgreen}{(+0.8)} \\ \hline
    Sup. pre-train+SPD & \cmark & \textbf{88.6} (+0.4) & \multicolumn{1}{l|}{\textbf{87.8} (+0.0)}  & \textbf{12.0} \textcolor{forestgreen}{(+0.6)} & \textbf{93.8} \textcolor{forestgreen}{(+0.7)}  & \textbf{97.5} (+0.1) & \multicolumn{1}{l|}{\textbf{94.6} (+0.1)}  & \textbf{36.3} \textcolor{forestgreen}{(+1.1)} & \textbf{94.6} (+0.2) \\ \hline 
    \end{tabular}
	}
\end{table*}

For the 1-class setting, the results of PaDiM with various pre-training options w/wo SPD are shown in Table \ref{tab:1cls}. The results are also visualized as scatter plots in Fig. \ref{fig:scatter}. We have several key observations. First, SPD improves performances of both anomaly detection and segmentation across almost all pre-training baselines on both VisA and MVTec-AD. While we report both AU-PR and AU-ROC, the former metric is more relevant to the application and we see that self-supervised methods are improved up to AU-PR of $2.6\%$. Note both metrics are averaged over the 12 objects in VisA. For different objects, the gains differ and are given in Sec. \textcolor{red}{D} of the supplementary. Second, the gap between self-supervised pre-training with SimSiam, SimCLR, MoCo, and supervised pre-training is large. SPD reduces this gap, but no combination of SSL and SPD beats supervised pre-training. This is in contrast to the low-shot regime in Section \ref{sect:few-shot-results}, where self-supervision has advantages in some cases. Third, PaDiM is one of the SOTA methods with $>97\%$ AU-ROC in MVTec. But it just achieves $<90\%$ AU-PR and AU-ROC in VisA - classification. For VisA - segmentation, PaDiM only achieves about $10\%$ AU-PR. This shows the difficulty of the VisA 1-class benchmark. Moreover, the gap between low AU-PR and high AU-ROC for both VisA/MVTec segmentation justifies the inflated performance view of AU-ROC, in favor of AU-PR as a more suitable metric in imbalanced datasets. In addition, even in terms of AU-ROC, the SPD consistently improves almost all baselines. 

\begin{table*}[!t]
	\centering
    \caption{2-class fine-tuning with different pre-training on VisA high-shot setup.}
    \label{tab:ava_2cls_high}
	\setlength{\tabcolsep}{7pt}
	\resizebox{0.75\linewidth}{!}{
	\centering
    \begin{tabular}{c|c|llll}
    \hline
    \multirow{2}{*}{} & \multirow{2}{*}{\begin{tabular}[c]{@{}c@{}}ImageNet\\ labels\end{tabular}} & \multicolumn{4}{c}{VisA (2-class, high-shot)} \\ \cline{3-6} 
                       &  & \multicolumn{2}{c|}{Classification} & \multicolumn{2}{c}{Segmentation} \\ \hline
                       &  & AU-PR & \multicolumn{1}{c|}{AU-ROC} & AU-PR          & AU-ROC          \\ \hline
    Sup. pre-train     & \cmark & 97.5  & \multicolumn{1}{l|}{99.5}  & 65.1           & 97.3            \\ \hline \hline
    SimSiam            & \xmark & 88.7  & \multicolumn{1}{l|}{97.9}   & 53.8           & 97.3            \\
    +SPD               & \xmark & 93.2 \textcolor{forestgreen}{(+4.5)} & \multicolumn{1}{l|}{98.7 \textcolor{forestgreen}{(+0.8)}}  & 59.7 \textcolor{forestgreen}{(+5.9)}          & 98.1 \textcolor{forestgreen}{(+0.8)}           \\ \hline
    MoCo               & \xmark & 93.9  & \multicolumn{1}{l|}{98.8}   & 62.4           & 98.0            \\
    +SPD               & \xmark & 94.2 (+0.3) & \multicolumn{1}{l|}{98.8 (+0.0)}   & 64.4 \textcolor{forestgreen}{(+2.0)}          & 97.9 (-0.1)           \\ \hline
    SimCLR             & \xmark & 93.4  & \multicolumn{1}{l|}{98.5}   & 67.7           & 95.3            \\
    +SPD               & \xmark & 92.7 (-0.7) & \multicolumn{1}{l|}{98.6 (+0.1)}   & 68.2 \textcolor{forestgreen}{(+0.5)}          & 95.7 (+0.4)           \\ \hline
    Sup. pre-train+SPD & \cmark & \textbf{98.3} \textcolor{forestgreen}{(+0.8)} & \multicolumn{1}{l|}{\textbf{99.7} (+0.2)}   & \textbf{71.9} \textcolor{forestgreen}{(+6.8)}          & \textbf{98.5} \textcolor{forestgreen}{(+1.2)}           \\ \hline
    \end{tabular}%
    }
\end{table*}

In Table \ref{tab:ava_2cls_high}, we show the results for the 2-class high-shot regime on the VisA and observe similar trends as above. However, the AU-PR gains from SPD on top of SimSiam and supervised pre-training are higher at 5.9\% and 6.8\% respectively for segmentation. Another key point to note here is that the AU-ROC metrics are saturating even though AU-PR metrics show room for improvement, particularly for segmentation. This another data point for preferring AU-PR metric. 
Comparing Tables \ref{tab:1cls} and \ref{tab:ava_2cls_high}, there is a significant gap between 1-class and 2-class performance on VisA. As anomalies are harder to obtain compared to normal images, bridging the gap is an open challenge to the research community.

\subsection{SPD in Low-shot 2-class Regime}\label{sect:few-shot-results}

\begin{table*}[!t]
	\centering
	\caption{Low-shot anomaly detection and segmentation on VisA.}
    \label{tab:low_shot}
	\setlength{\tabcolsep}{7pt}
	\resizebox{\linewidth}{!}{
	\centering
    \begin{tabular}{c|c|llll||llll}
    \hline
    \multirow{2}{*}{}  & \multirow{2}{*}{\begin{tabular}[c]{@{}c@{}}ImageNet\\ labels\end{tabular}} & \multicolumn{4}{c||}{Classification (2-class, low-shot)}                                & \multicolumn{4}{c}{Segmentation (2-class, low-shot)}                                  \\ \cline{3-10} 
                       &                                                                            & \multicolumn{2}{c|}{5-shot}                    & \multicolumn{2}{c||}{10-shot} & \multicolumn{2}{c|}{5-shot}                    & \multicolumn{2}{c}{10-shot} \\ \hline
                       &                                                                            & AU-PR       & \multicolumn{1}{c|}{AU-ROC}      & AU-PR         & AU-ROC       & AU-PR       & \multicolumn{1}{c|}{AU-ROC}      & AU-PR        & AU-ROC       \\ \hline
    Sup. pre-train     &   \cmark                                                                         & 59.2        & \multicolumn{1}{l|}{85.5}        & 70.4          & 91.7         & 17.8        & \multicolumn{1}{l|}{74.6}        & 28.3         & 81.8         \\ \hline\hline
    SimSiam            &   \xmark                                                                          & 51.9        & \multicolumn{1}{l|}{82.3}        & 65.0          & 89.4         & 17.3        & \multicolumn{1}{l|}{75.2}        & 28.5         & 81.6         \\
    +SPD               &   \xmark                                                                        & 56.1 \textcolor{forestgreen}{(+4.2)} & \multicolumn{1}{c|}{84.0 \textcolor{forestgreen}{(+1.7)}} & 67.6 \textcolor{forestgreen}{(+2.6)}   & 90.8 \textcolor{forestgreen}{(+1.4)}  & 18.2 \textcolor{forestgreen}{(+0.9)} & \multicolumn{1}{l|}{\textbf{76.0} \textcolor{forestgreen}{(+0.8)}} & 29.7 \textcolor{forestgreen}{(+1.2)}  & 83.2 \textcolor{forestgreen}{(+1.6)}  \\ \hline
    MoCo               &   \xmark                                                                         & 56.1        & \multicolumn{1}{l|}{83.8}        & 68.7          & 90.6         & 21.5        & \multicolumn{1}{l|}{80.5}        & 32.3         & \textbf{85.7}         \\
    +SPD               &   \xmark                                                                         & 56.4 (+0.3) & \multicolumn{1}{l|}{83.9 (+0.1)} & 68.0 (-0.7)   & 90.1 (-0.5)  & \textbf{22.1} \textcolor{forestgreen}{(+0.6)} & \multicolumn{1}{l|}{78.5 (-2.0)} & \textbf{32.8} \textcolor{forestgreen}{(+0.5)}  & 84.9 (-0.8)  \\ \hline 
    SimCLR             &  \xmark                                                                          & 48.4        & \multicolumn{1}{l|}{79.6}        & 58.2          & 86.0         & 18.4        & \multicolumn{1}{l|}{71.2}        & 23.0         & 75.1         \\
    +SPD               &  \xmark                                                                         & 47.4 (-1.0) & \multicolumn{1}{l|}{79.9 (+0.3)} & 59.0 \textcolor{forestgreen}{(+0.8)}   & 86.1 (+0.1)  & 18.9 \textcolor{forestgreen}{(+0.5)} & \multicolumn{1}{c|}{74.5 \textcolor{forestgreen}{(+3.3)}} & 25.1 \textcolor{forestgreen}{(+2.1)}  & 78.2 \textcolor{forestgreen}{(+3.1)}  \\ \hline
    Sup. pre-train+SPD &  \cmark                                                                          &    \textbf{59.8} \textcolor{forestgreen}{(+0.6)}         & \multicolumn{1}{l|}{\textbf{85.9} (+0.4)}            &         \textbf{71.2} \textcolor{forestgreen}{(+0.8)}      &   \textbf{92.1} (+0.4)          & 18.7 \textcolor{forestgreen}{(+0.9)} & \multicolumn{1}{c|}{75.9 \textcolor{forestgreen}{(+1.3)}} & 30.6 \textcolor{forestgreen}{(+2.3)}  & 81.8 (+0.0)  \\ \hline
    \end{tabular}%
    }
\end{table*}

\noindent\textbf{Low-shot anomaly segmentation:} With different ImageNet pre-training as initialization, a 2-class U-Net with ResNet-50 encoder is trained for each 5/10-shot segmentation setup. From Table \ref{tab:low_shot}, SPD again improves all baselines in both 5-shot and 10-shot evaluation, with AU-PR gain up to $2.3\%$. One departure from the high-shot regime is that for few-shot anomaly segmentation, MoCo+SPD is the best method, even outperforming supervised pre-training.   

\noindent\textbf{Low-shot anomaly detection:} Initialized with different ImageNet pre-training, a 2-class ResNet-50 is trained in 5/10-shot setups for anomaly detection. From Table \ref{tab:low_shot}, overall the supervised pre-training with SPD outperforms both supervised pre-training only and other SSL's. Moreover, SPD significantly improves SimSiam with $4.2\%$ AU-PR in 5-shot and $2.6\%$ AU-PR in 10-shot, although it's still inferior to supervised pre-training.
\subsection{Ablation Study}

\begin{table*}[!t]
	\centering
	\caption{Ablation study}
	\label{tab:ablation}
	\setlength{\tabcolsep}{7pt}
	\resizebox{\linewidth}{!}{
	\centering
	\begin{tabular}{c|cccc|cccc}
    \hline
    \multirow{2}{*}{} & \multicolumn{4}{c|}{VisA (1-class)}                             & \multicolumn{4}{c}{MVTec-AD (1-class)}                         \\ \cline{2-9} 
     & \multicolumn{2}{c|}{Classification} & \multicolumn{2}{c|}{Segmentation} & \multicolumn{2}{c|}{Classification} & \multicolumn{2}{c}{Segmentation} \\ \hline
                      & AU-PR & \multicolumn{1}{c|}{AU-ROC} & AU-PR & AU-ROC & AU-PR & \multicolumn{1}{c|}{AU-ROC} & AU-PR & AU-ROC \\ \hline
    SimSiam w/ Res50     & 80.2 & \multicolumn{1}{c|}{78.1}   & 9.1  & 93.1  & 92.6 & \multicolumn{1}{c|}{83.9}  & 29.7 & 92.1  \\ \hline\hline
    +SPD $(\eta=0.1)$         & 82.8 & \multicolumn{1}{c|}{81.2}  & 9.4  & 92.7  & 94.1 & \multicolumn{1}{c|}{88.0}  & 32.0 & 92.2  \\ 
    +SPD $(\eta=0.5)$         & 80.5 & \multicolumn{1}{c|}{79.3}  & 8.7  & 93.0  & 93.3 & \multicolumn{1}{c|}{84.9}  & 30.1 & 91.9  \\
    +SPD $(\eta=1.0)$           & 81.5 & \multicolumn{1}{c|}{79.8}  & 9.4  & 92.8  & 93.4 & \multicolumn{1}{c|}{85.8}  & 30.0 & 92.0  \\ \hline
    +SPD w/ CutPaste    & 78.8 & \multicolumn{1}{c|}{77.0}  & 9.7  & 93.1  & 93.5 & \multicolumn{1}{c|}{85.2}  & 28.2 & 91.3  \\
    +SPD w/ Xent        & 71.4 & \multicolumn{1}{c|}{66.6}  & 2.7  & 84.8  & 86.3 & \multicolumn{1}{c|}{71.0}  & 15.2  & 82.6  \\ \hline
    SimSiam w/ WideRes50 & 80.3 & \multicolumn{1}{c|}{77.7}  & 9.9  & 93.6  & 93.0 & \multicolumn{1}{c|}{84.7}  & 31.3 & 92.2  \\
    +SPD              & 81.9 & \multicolumn{1}{c|}{80.4}  & 10.5  & 93.7  & 93.4 & \multicolumn{1}{c|}{85.4}  & 32.5 & 92.8  \\ \hline
    \end{tabular}
    }

\end{table*}
\begin{table*}[!t]
	\centering
    \caption{1-class performance evaluation on VisA and MVTec-AD with PatchCore.}
    \label{tab:1cls_patchcore}
	\setlength{\tabcolsep}{7pt}
	\resizebox{\linewidth}{!}{
	\centering
    \begin{tabular}{c|cccc|cccc}
    \hline
    \multirow{2}{*}{\begin{tabular}[c]{@{}c@{}}Backbone:\\ Wide ResNet50\end{tabular}} & \multicolumn{4}{c|}{VisA (1-class)}                                     & \multicolumn{4}{c}{MVTec-AD (1-class)}                                 \\ \cline{2-9} 
                                                                                   & \multicolumn{2}{c|}{Classification} & \multicolumn{2}{c|}{Segmentation} & \multicolumn{2}{c|}{Classification} & \multicolumn{2}{c}{Segmentation} \\ \hline
                          & AU-PR & \multicolumn{1}{c|}{AU-ROC} & AU-PR           & AU-ROC          & AU-PR & \multicolumn{1}{c|}{AU-ROC} & AU-PR          & AU-ROC          \\ \hline
    Sup. pre-train        & 93.3  & \multicolumn{1}{c|}{92.4}   & 38.4            & 98.4            & 99.2  & \multicolumn{1}{c|}{99.8}   & 48.8           & 97.6            \\
    Sup. pre-train+SPD    & 93.8 \textcolor{forestgreen}{(+0.5)} & \multicolumn{1}{c|}{92.5 (+0.1)}   & 39.3 \textcolor{forestgreen}{(+0.9)}           & 98.1 (-0.3)           & 99.0 (-0.2) & \multicolumn{1}{c|}{99.7 (-0.1)}   & 49.3 \textcolor{forestgreen}{(+0.5)}          & 97.5 (-0.1)           \\ \hline
    \end{tabular}%
	}
\end{table*}

We conduct extensive ablation studies based on ImageNet SimSiam pre-training and PaDiM as the anomaly detection and segmentation algorithms trained in the 1-class setups of VisA and MVTec-AD. Results are shown in Table \ref{tab:ablation}.

\noindent\textbf{Sensitivity analysis on SPD loss weight $\eta$:} From Table \ref{tab:ablation}, we see consistent improvement for $\eta=0.1,0.5,1.0$ in at least one task for both datasets. SPD loss with $\eta=0.1$ gives us the best performances in both datasets, which is chosen as the default SPD loss weight for all pre-training with SPD. So the SimSiam+SPD ($\eta=0.1$) is regarded as SimSiam+SPD for better clarity.

\noindent\textbf{Comparison between SPD and CutPaste \cite{li2021cutpaste}:} CutPaste and cross-entropy loss used in \cite{li2021cutpaste} for anomaly detection training can also be used in ImageNet pre-training. An ablation study is done to demonstrate the superiority of the proposed SmoothBlend and SPD loss. With $\mathcal{L}_{\mathrm{SPD}}$, SmoothBlend is arguably better than CutPaste by $4.0\%$ and $3.8\%$ AU-PR improvement in VisA - classification and MVTec - segmentation (+SPD v.s. +SPD w/ CutPaste). With the SmoothBlend, the SPD loss significantly outperforms cross-entropy loss (+SPD v.s. +SPD w/ Xent). Such results demonstrate the validity of proposed methods.

\noindent\textbf{SPD with different backbones:} ResNet-50 is adopted as the backbone for all major experiments in this paper. We demonstrate the SPD can generalize to different network architectures by experiments of SimSiam w/wo SPD on wide ResNet-50 \cite{zagoruyko2016wide}. As in Table \ref{tab:ablation}, SPD still improves the baseline.

\noindent\textbf{Results with PatchCore: }
In addition to PaDiM, we also evaluate supervised pre-trained models based on another state-of-the-art 1-class method PatchCore \cite{Roth_2022_CVPR}. Wide ResNet-50 is chosen as the backbone network. As in Table \ref{tab:1cls_patchcore}, on VisA, SPD improves supervised pre-trained model by $0.5\%$ and $0.9\%$ AU-PR for both classification and segmentation. On MVTec-AD, SPD improves by $0.5\%$ AU-PR for segmentation with slightly performance decreased in classification.

\noindent\textbf{Extending SPD to other tasks:} Besides improvement on defect detection and segmentation, SPD also improves ImageNet supervised classification accuracy:  $69.8\%\rightarrow70.2\%$ for ResNet-18 and $76.1\%\rightarrow76.4\%$ for ResNet-50.
Pre-trained models with better ImageNet accuracy are expected to benefit downstream tasks more. Thus we speculate that SPD will work well for object recognition and detection, especially on fine-grained classification and small object detection as SPD promotes local sensitivity. In addition, we will leverage the proposed SPD training as a 1-class anomaly detection model to be trained by downstream data.

\noindent\textbf{Qualitative results:} To qualitatively demonstrate the effectiveness of SPD regularization, we represent attention maps and anomaly segmentation in Sec. \textcolor{red}{E} of the supplementary due to page limits.

\section{Conclusions}
In this work, we present a spot-the-difference (SPD) training to regularize pre-trained models' local sensitivity to anomalous patterns. We also present a novel Visual Anomaly (VisA) dataset which is the largest industrial anomaly detection dataset. Extensive experiments demonstrate the benefits of SPD for various contrastive self-supervised and supervised pre-training for anomaly detection and segmentation. Compared to standard supervised pre-training, SimSiam with SPD obtains superior or competitive performances in low-shot regime while supervised learning with SPD presents better performances in various setups. 

\section*{Acknowledgments}
The authors would like to thank Fanyi Xiao, Erhan Bas, Aditya Deshpande and Joachim Stahl for idea brainstorming
and providing insightful comments on the manuscript.

\bibliographystyle{splncs04}
\bibliography{spot-diff}

\clearpage

\section*{Appendix}
In this appendix, we present the additional details and
results that are not covered by the main paper.

\appendix

\section{Scatter Plots for AU-ROC}
\begin{figure}[!ht]
\centering
\includegraphics[width=0.45\textwidth]{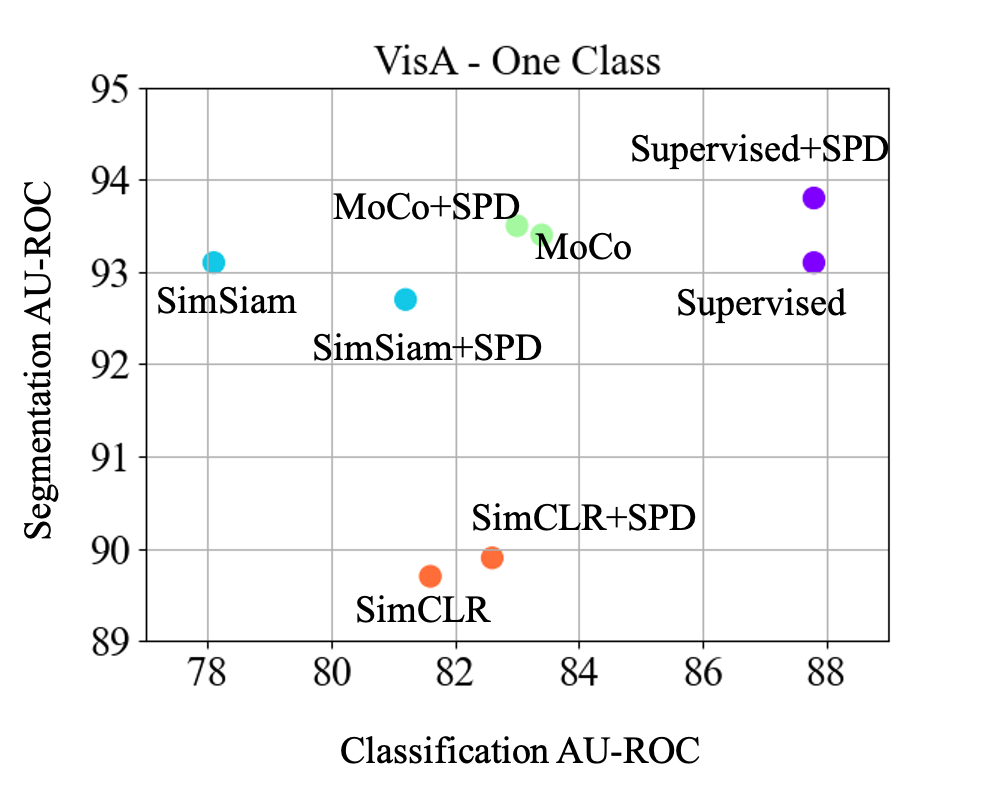}
\includegraphics[width=0.45\textwidth]{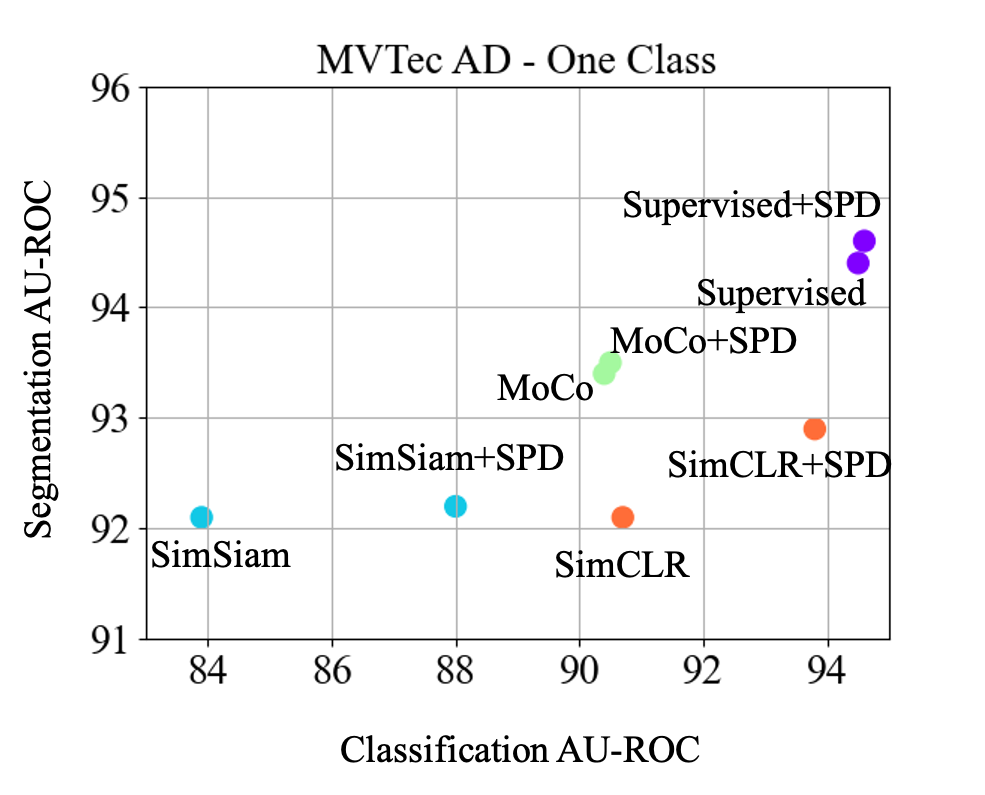}\\
 \caption{Comparison of classification \textit{vs.} segmentation AU-ROCs for various ImageNet pre-trained representations in the 1-class setup.}
 \label{fig:scatter_auroc}
\end{figure}

The scatter plots in terms of AU-PR for different ImageNet pre-trained models on 1-class training setups of VisA and MVTec-AD are shown in Fig. \ref{fig:scatter} of the main paper. In addition, we also show the scatter plots in terms of AU-ROC in Fig. \ref{fig:scatter_auroc}. The SPD almost improves AU-ROC for all baselines on both classification and segmentation tasks of VisA and MVTec AD, demonstrating the effectiveness of the proposed SPD training.

\section{Further Discussion on AU-PR and AU-ROC}
AU-ROC is a good metric for balanced dataset. However, as mentioned in several past works \cite{cook2020consult,davis2006relationship,saito2015precision}, in imbalanced dataset where minor class is more important, AU-ROC provides an inflated view of performance about the minor class and AU-PR is more informative. Imbalance is common in anomaly detection/segmentation datasets. Most experimental results in this work and \cite{bergmann2021mvtec} demonstrate the point. For example, considering the results of 1-cls segmentation on VisA, even when a model achieves $>95\%$ AU-ROC, the AU-PR can be $<10\%$. Moreover, the AU-ROC can be misleading about the performance on minor class in imbalanced dataset. Specifically, a model with a lower AU-ROC might be better than another model with higher AU-ROC in terms of the performance on anomaly class (reflected by AU-PR), although it might be worse in the major class. To give more intuition, we present the following toy example to demonstrate the above points. 

\noindent\textbf{A toy example: } 
First, we denote $P$ as ground truth positives, N as ground truth negatives, $TP$ as True Positive, $FP$ as False Positive, $FN$ as False Negative, $TN$ as True Negative. Then we define the following metrics.

\begin{equation}
    Precision=\frac{TP}{TP+FP}
    \label{eqn:prec}
\end{equation}

\begin{equation}
    Recall=\frac{TP}{P}=\frac{TP}{TP+FN}
    \label{eqn:recall}
\end{equation}

\begin{equation}
    FPR=\frac{FP}{N}=\frac{FP}{TN+FP}=\frac{N-TN}{N}
    \label{eqn:FPR}
\end{equation}

\begin{equation}
    F_1=\frac{2*Precision*Recall}{Precision+Recall}=\frac{TP}{TP+0.5*(FP+FN)}
    \label{eqn:F1}
\end{equation}

\begin{figure}[!ht]
\centering
\includegraphics[width=0.45\textwidth]{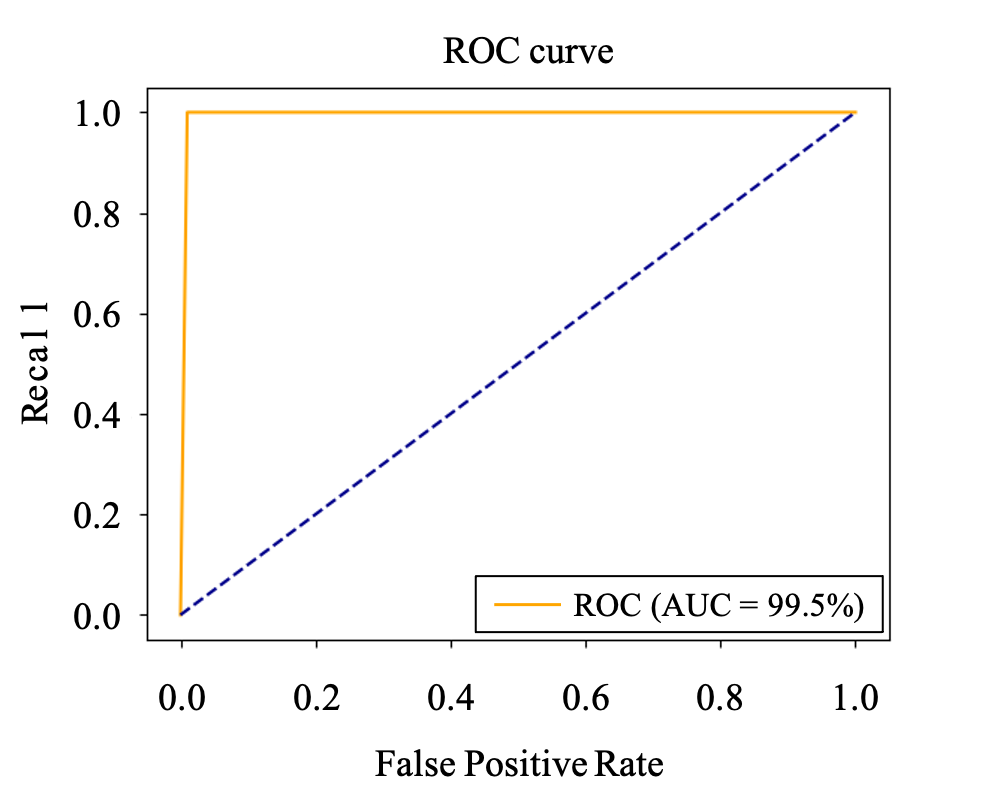}
\includegraphics[width=0.45\textwidth]{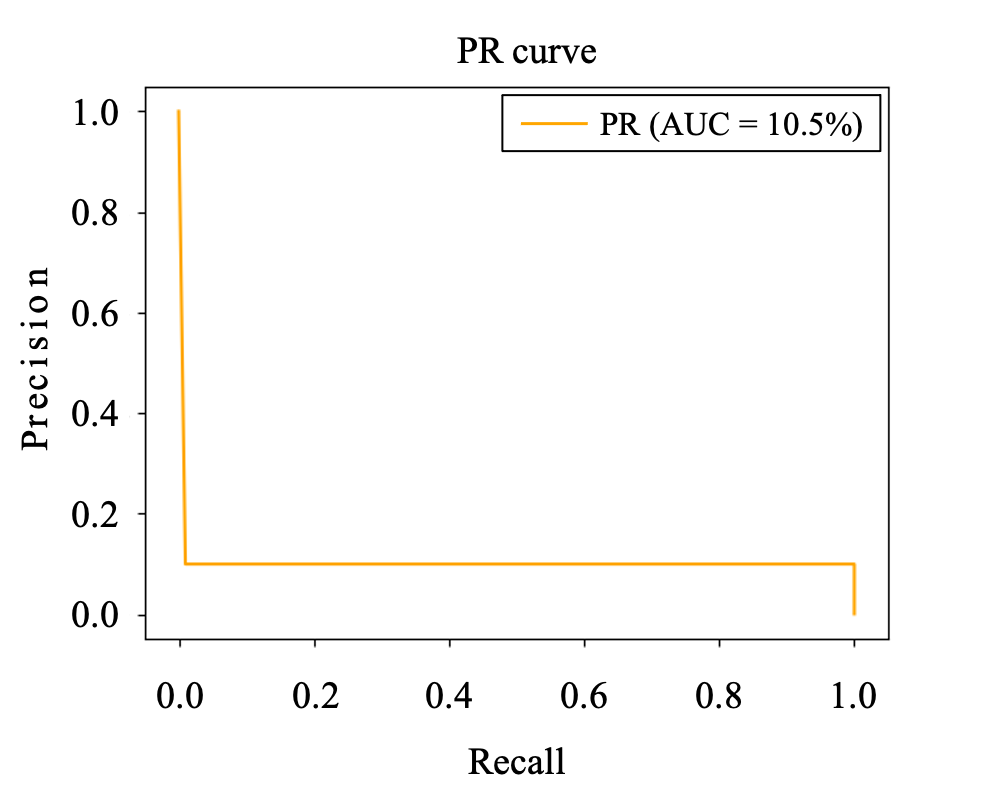}\\
 \caption{ROC and PR curves for model A}
 \label{fig:curves_modelA}
\end{figure}

\begin{figure}[!ht]
\centering
\includegraphics[width=0.45\textwidth]{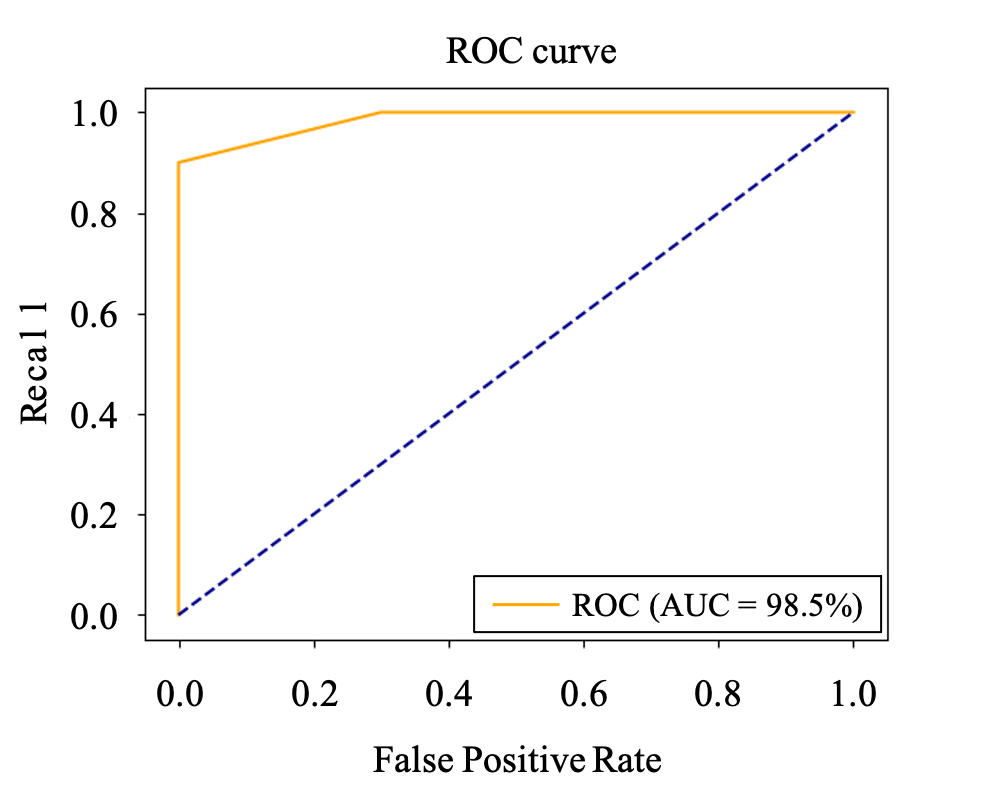}
\includegraphics[width=0.45\textwidth]{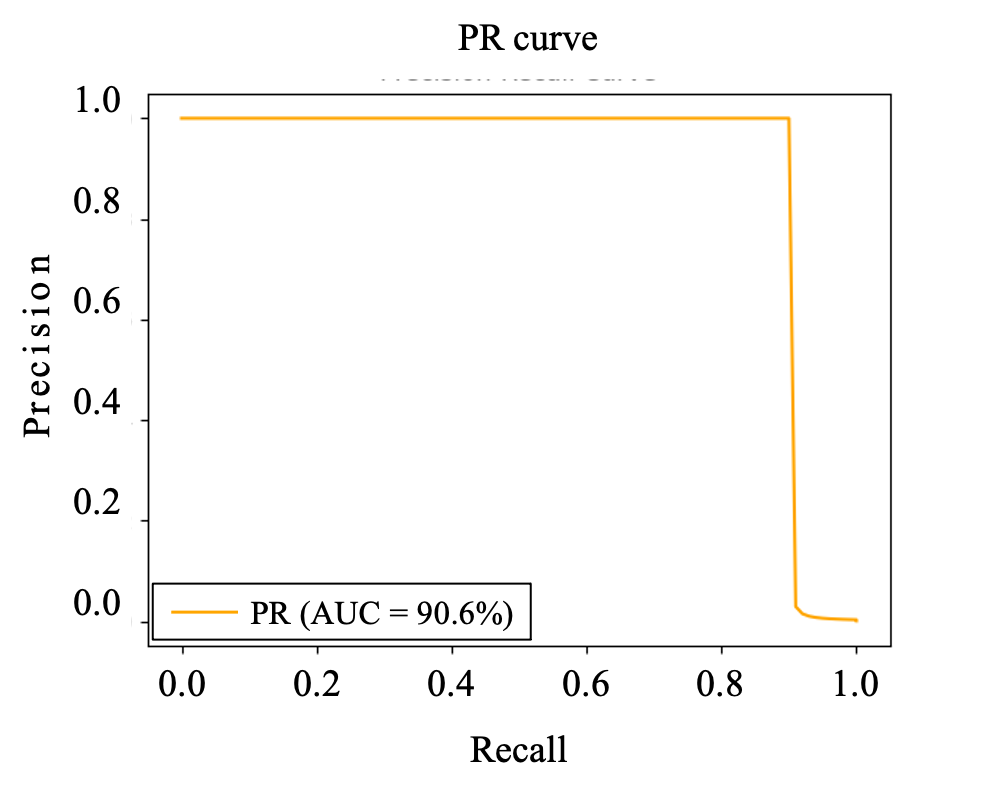}\\
 \caption{ROC and PR curves for model B}
 \label{fig:curves_modelB}
\end{figure}

Varying the operating thresholds, ROC curve measures the trade-off between recall and FPR and AU-ROC is the Area Under the ROC curve. PR curve measures trade-off between precision and recall and AU-PR is the Area Under the PR curve. Max $F_1$ is the best $F_1$ that can be obtained from the PR curve.

Then we define a toy test set with 100 gt positives (anomaly), and gt 100,000 negatives (normal). This test set is imbalanced with neg/pos ratio = 1,000. Note that the scores of gt positive samples are the thresholds deciding the shapes of ROC and PR curves. In the following, we define model A and model B.

Model A has the following behaviors on the test set. To correctly predict each TP when threshold reduces, additional 10 FPs will be produced, leading to (TP, FP) pairs (1, 10), (2, 20), ..., (100, 1,000). We plot corresponding ROC and PR curves in Fig. \ref{fig:curves_modelA}. The AU-ROC=99.5\% which seems to indicate the model is close to perfection. However, the AU-PR is just 10.5\% and the Max $F_1$ is 18.2\%. At the best threshold, model A only has 10\% precision with 100\% recall. Although the AU-ROC is inflatedly high, model A has a poor performance in predicting the positive class. 

For model B, to correctly predict the first 90 TPs, there are no FPs. But in the remaining 10 positive samples, 3,000 FPs will be produced for each TP. So the (TP, FP) pairs are (1, 0), (2, 0), ..., (90, 0), (91, 3,000), (92, 6,000),...,(100, 30,000). We plot corresponding ROC and PR curves in Fig. \ref{fig:curves_modelB}. 

Comparing with model A, model B has a worse AU-ROC (98.5\% v.s. 99.5\%). However, comparing at the best operating point, model B is much better than model A in predicting positive samples and achieves 100\% precision and 90\$ recall (v.s. 10\% precision and 100\% recall). Model B reaches 90.6\% AU-PR and 94.7\% Max $F_1$ which are much better than model A’s 10.5\% AU-PR and 18.2\% Max $F_1$. In such case, the AU-ROC provides inflated and misleading view about model performance in positive predictions.

\section{Implementation Details}
\noindent\textbf{Pre-training:} First, we set the SPD loss weight $\eta=0.1$ for all the experiments unless specified otherwise. Second, for each baseline (SimSiam \cite{chen2021exploring}, MoCo \cite{he2020momentum}, SimCLR \cite{chen2020simple}) with SPD, we follow exactly the same default hyperparamters in the baseline. Third, for SmoothBlend, the area of the cut patch is $0.5\% - 1\%$ in relative to the full image's size. The cut patch's aspect ratio ranges from $0.3$ to $3$. The standard deviations of kernel in Gaussian smoothing applied to the $\alpha$ mask are $(8,8)$. Each image only has one smoothly blended patch. Fourth, for weak global augmentations, we choose random cropping with a ratio $[0.9,1.0]$, color jittering with brightness$=0.1$, contrast$=0.1$, saturation$=0.1$, hue$=0.05$, Gaussian blur with standard deviations $(0.1, 0.3)$, horizontal flipping. Note that the color jittering and Gaussian blur are applied randomly with $0.8$ and $0.5$ probability.

\noindent\textbf{Downstream anomaly detection and segmentation models:} For PaDiM, we follow exactly the same hyperparameters in \cite{defard2021padim}. For two-class supervised networks, in high-shot setups, we fine-tune the models for 80 epochs with SGD with learneable backbone parameters. In few-shot setups, we train the models for $1,000$ iterations for few-shot classification and $500$ iterations for few-shot segmentation. We choose a fixed learning rate policy with $lr=0.0001$.

\section{Full results for each subset of VisA}
In this section, we present the results for each subset of VisA w.r.t. SimSiam, SimSiam+SPD, supervised and supervised+SPD pre-training. Tables \ref{tab:1cls_cls_full} and \ref{tab:1cls_seg_full} provide the results for 1-class classification and segmentation training setups with PaDiM. Tables \ref{tab:2cls_cls_full} and \ref{tab:2cls_seg_full} present the results for 2-class high-shot classification and segmentation training setups. Tables \ref{tab:2cls_cls_5shot_full} and \ref{tab:2cls_seg_5shot_full} show the results for 2-class 5-shot classification and segmentation training setups. Tables \ref{tab:2cls_cls_10shot_full} and \ref{tab:2cls_seg_10shot_full} give the results for 2-class 10-shot classification and segmentation training setups. Generally speaking, the VisA subsets with multiple instances (Macaroni1, Macaroni2, Capsules, Candles) are the most difficult cases with lowest scores. The VisA subsets with complex structures (PCB1, PCB2, PCB3, PCB4) are relatively easier than the multiple instances cases with better scores. The VisA subsets with single instance (Cashew, Chewing gum, Fryum, Pipe Fryum) are easier than the complex structure cases.
\begin{table*}[!ht]
	\centering
	\caption{1-class anomaly detection on VisA.}
    \label{tab:1cls_cls_full}
    \vspace{-2mm}
	\setlength{\tabcolsep}{7pt}
	\resizebox{\linewidth}{!}{
	\centering
\begin{tabular}{cc|cccc|cccc}
\hline
                                                         &             & \multicolumn{2}{c|}{SimSiam}        & \multicolumn{2}{c|}{+SPD} & \multicolumn{2}{c|}{Supervised}     & \multicolumn{2}{c}{+SPD} \\ \hline
\multicolumn{1}{c|}{}                                    &             & AU-PR & \multicolumn{1}{c|}{AU-ROC} & AU-PR       & AU-ROC      & AU-PR & \multicolumn{1}{c|}{AU-ROC} & AU-PR      & AU-ROC      \\ \hline
\multicolumn{1}{c|}{\multirow{4}{*}{Complex structure}}  & PCB1        & 83.5  & 85.4                        & 83.5        & 86.8       & 89.5  & 92.0                        & 90.4       & 92.7        \\
\multicolumn{1}{c|}{}                                    & PCB2        & 76.1  & 76.6                        & 76.9        & 76.6        & 88.5  & 89.1                        & 87.0       & 87.9        \\
\multicolumn{1}{c|}{}                                    & PCB3        & 73.0  & 75.0                        & 72.0        & 72.2        & 87.8  & 87.8                        & 86.9       & 85.4        \\
\multicolumn{1}{c|}{}                                    & PCB4        & 92.3  & 93.9                        & 93.8        & 95.2        & 98.3  & 98.5                        & 98.9       & 99.1        \\ \hline
\multicolumn{1}{c|}{\multirow{4}{*}{Multiple instances}} & Macaroni1   & 67.7  & 72.2                        & 74.4        & 75.7        & 81.3  & 84.8                        & 82.2       & 85.7        \\
\multicolumn{1}{c|}{}                                    & Macaroni2   & 56.1  & 59.2                        & 62.3        & 66.8        & 63.6  & 69.8                        & 66.7       & 70.8        \\
\multicolumn{1}{c|}{}                                    & Capsules    & 70.4  & 58.1                        & 72.5        & 62.0        & 74.8  & 65.8                        & 78.6       & 68.1        \\
\multicolumn{1}{c|}{}                                    & Candles     & 81.1  & 83.9                        & 83.7        & 85.3        & 86.5  & 88.9                        & 86.2       & 89.1        \\ \hline
\multicolumn{1}{c|}{\multirow{4}{*}{Single instance}}    & Cashew      & 90.5  & 79.7                        & 93.8        & 86.6        & 95.7  & 90.6                        & 95.8       & 90.5        \\
\multicolumn{1}{c|}{}                                    & Chewing gum & 92.8  & 90.1                        & 98.2        & 96.7        & 99.6  & 99.2                        & 99.7       & 99.3        \\
\multicolumn{1}{c|}{}                                    & Fryum       & 89.0  & 81.5                        & 89.4        & 83.6        & 94.2  & 89.8                        & 93.7       & 89.8        \\
\multicolumn{1}{c|}{}                                    & Pipe fryum  & 89.9  & 81.6                        & 93.6        & 87.1        & 98.5  & 97.2                        & 97.6       & 95.6        \\ \hline
\multicolumn{1}{c|}{}                                    & Mean  & 80.2 & 78.1 & 82.84 & 81.2 & 88.2 & 7.80 & 88.6 & 87.8 \\ \hline
\end{tabular}%
    }
\end{table*}

\begin{table*}[!ht]
	\centering
	\caption{1-class anomaly segmentation on VisA.}
    \label{tab:1cls_seg_full}
    \vspace{-2mm}
	\setlength{\tabcolsep}{7pt}
	\resizebox{\linewidth}{!}{
	\centering
    \begin{tabular}{cc|cccccccc}
    \hline
                                                             &             & \multicolumn{2}{c|}{SimSiam}        & \multicolumn{2}{c|}{+SPD}           & \multicolumn{2}{c|}{Supervised}     & \multicolumn{2}{c}{+SPD} \\ \hline
    \multicolumn{1}{c|}{}                                    &             & AU-PR & \multicolumn{1}{c|}{AU-ROC} & AU-PR & \multicolumn{1}{c|}{AU-ROC} & AU-PR & \multicolumn{1}{c|}{AU-ROC} & AU-PR      & AU-ROC      \\ \hline
    \multicolumn{1}{c|}{\multirow{4}{*}{Complex structure}}  & PCB1        & 9.8   & 94.8                        & 13.1  & \multicolumn{1}{c|}{96.4}   & 18.2  & 97.3                        & 21.1       & 97.7        \\
    \multicolumn{1}{c|}{}                                    & PCB2        & 9.8   & 96.3                        & 9.6   & \multicolumn{1}{c|}{96.3}   & 12.4  & 97.4                        & 11.8       & 97.2        \\
    \multicolumn{1}{c|}{}                                    & PCB3        & 10.2  & 96.0                        & 10.5  & \multicolumn{1}{c|}{96.2}   & 14.8  & 96.1                        & 15.7       & 96.7        \\
    \multicolumn{1}{c|}{}                                    & PCB4        & 6.4   & 88.2                        & 8.5   & \multicolumn{1}{c|}{86.7}   & 12.0  & 88.4                        & 11.0       & 89.2        \\ \hline
    \multicolumn{1}{c|}{\multirow{4}{*}{Multiple instances}} & Macaroni1   & 2.9   & 97.9                        & 3.4   & \multicolumn{1}{c|}{97.7}   & 5.0   & 98.8                        & 4.0        & 98.8        \\
    \multicolumn{1}{c|}{}                                    & Macaroni2   & 0.3   & 93.9                        & 0.6   & \multicolumn{1}{c|}{94.3}   & 0.8   & 95.6                        & 1.0        & 96.0        \\
    \multicolumn{1}{c|}{}                                    & Capsules    & 1.3   & 84.3                        & 2.7   & \multicolumn{1}{c|}{87.5}   & 2.8   & 83.8                        & 3.2        & 86.3        \\
    \multicolumn{1}{c|}{}                                    & Candles     & 3.5   & 95.6                        & 3.5   & \multicolumn{1}{c|}{93.7}   & 6.7   & 96.8                        & 7.3        & 97.3        \\ \hline
    \multicolumn{1}{c|}{\multirow{4}{*}{Single instance}}    & Cashew      & 9.9   & 88.8                        & 9.5   & \multicolumn{1}{c|}{86.3}   & 10.0  & 85.5                        & 8.7        & 86.1        \\
    \multicolumn{1}{c|}{}                                    & Chewing gum & 29.7  & 97.3                        & 28.5  & \multicolumn{1}{c|}{97.0}   & 29.2  & 96.1                        & 31.3       & 96.9        \\
    \multicolumn{1}{c|}{}                                    & Fryum       & 11.6  & 90.1                        & 11.7  & \multicolumn{1}{c|}{89.0}   & 12.1  & 88.2                        & 11.9       & 88.0        \\
    \multicolumn{1}{c|}{}                                    & Pipe fryum  & 13.2  & 94.4                        & 11.7  & \multicolumn{1}{c|}{91.6}   & 12.5  & 93.3                        & 16.7       & 95.4        \\ \hline
    \multicolumn{1}{c|}{}                                    & Mean        &       9.1 & 93.1 & 9.4 & \multicolumn{1}{c|}{92.7} & 11.4 & 93.1 & 12.0 & 93.8 \\ \hline
    \end{tabular}%
    }
\end{table*}

\begin{table*}[!ht]
	\centering
	\caption{2-class high-shot anomaly detection on VisA.}
    \label{tab:2cls_cls_full}
    \vspace{-2mm}
	\setlength{\tabcolsep}{7pt}
	\resizebox{\linewidth}{!}{
	\centering
    \begin{tabular}{cc|cccc|cccc}
    \hline
                                                             &             & \multicolumn{2}{c|}{SimSiam}        & \multicolumn{2}{c|}{+SPD} & \multicolumn{2}{c|}{Supervised}     & \multicolumn{2}{c}{+SPD} \\ \hline
    \multicolumn{1}{c|}{}                                    &             & AU-PR & \multicolumn{1}{c|}{AU-ROC} & AU-PR       & AU-ROC      & AU-PR & \multicolumn{1}{c|}{AU-ROC} & AU-PR      & AU-ROC      \\ \hline
    \multicolumn{1}{c|}{\multirow{4}{*}{Complex structure}}  & PCB1        & 79.8  & 96.7                        & 84.9        & 98.4        & 89.9  & 98.8                        & 93.4       & 99.4        \\
    \multicolumn{1}{c|}{}                                    & PCB2        & 95.9  & 98.8                        & 96.9        & 99.3        & 98.2  & 99.7                        & 96.9       & 99.3        \\
    \multicolumn{1}{c|}{}                                    & PCB3        & 86.0  & 98.0                        & 94.0        & 99.1        & 99.8  & 100.0                       & 99.4       & 99.9        \\
    \multicolumn{1}{c|}{}                                    & PCB4        & 98.7  & 99.8                        & 99.7        & 100.0       & 99.1  & 99.9                        & 99.9       & 100.0       \\ \hline
    \multicolumn{1}{c|}{\multirow{4}{*}{Multiple instances}} & Macaroni1   & 88.2  & 98.5                        & 95.0        & 99.3        & 99.5  & 99.9                        & 99.9       & 100.0       \\
    \multicolumn{1}{c|}{}                                    & Macaroni2   & 82.4  & 97.2                        & 93.2        & 98.8        & 99.9  & 100.0                       & 100.0      & 100.0       \\
    \multicolumn{1}{c|}{}                                    & Capsules    & 70.4  & 91.6                        & 76.6        & 93.7        & 88.4  & 97.0                        & 94.3       & 98.6        \\
    \multicolumn{1}{c|}{}                                    & Candles     & 89.5  & 98.4                        & 89.8        & 98.1        & 97.8  & 99.6                        & 98.2       & 99.7        \\ \hline
    \multicolumn{1}{c|}{\multirow{4}{*}{Single instance}}    & Cashew      & 80.0  & 97.0                        & 92.7        & 98.7        & 99.1  & 99.8                        & 98.4       & 99.7        \\
    \multicolumn{1}{c|}{}                                    & Chewing gum & 98.1  & 99.4                        & 98.4        & 99.5        & 99.6  & 99.8                        & 100.0      & 100.0       \\
    \multicolumn{1}{c|}{}                                    & Fryum       & 99.5  & 99.9                        & 99.6        & 99.9        & 99.5  & 99.9                        & 99.6       & 99.9        \\
    \multicolumn{1}{c|}{}                                    & Pipe fryum  & 95.8  & 99.5                        & 97.7        & 99.7        & 99.2  & 99.9                        & 99.3       & 99.9        \\ \hline
    \multicolumn{1}{c|}{}                                    & Mean        &  88.7 & 97.9 & 93.2 & 98.7 & 97.5 & 99.5 & 98.3 & 99.7         \\ \hline
    \end{tabular}%
    }
\end{table*}

\begin{table*}[!ht]
	\centering
	\caption{2-class high-shot anomaly segmentation on VisA.}
    \label{tab:2cls_seg_full}
    \vspace{-2mm}
	\setlength{\tabcolsep}{7pt}
	\resizebox{\linewidth}{!}{
	\centering
    \begin{tabular}{cc|cccc|cccc}
    \hline
                                                             &             & \multicolumn{2}{c|}{SimSiam}        & \multicolumn{2}{c|}{+SPD} & \multicolumn{2}{c|}{Supervised}     & \multicolumn{2}{c}{+SPD} \\ \hline
    \multicolumn{1}{c|}{}                                    &             & AU-PR & \multicolumn{1}{c|}{AU-ROC} & AU-PR       & AU-ROC      & AU-PR & \multicolumn{1}{c|}{AU-ROC} & AU-PR      & AU-ROC      \\ \hline
    \multicolumn{1}{c|}{\multirow{4}{*}{Complex structure}}  & PCB1        & 79.6  & 98.6                        & 85.1        & 99.4        & 59.4  & 93.6                        & 91.3       & 99.2        \\
    \multicolumn{1}{c|}{}                                    & PCB2        & 32.2  & 95.2                        & 31.5        & 95.4        & 52.1  & 98.2                        & 66.6       & 98.4        \\
    \multicolumn{1}{c|}{}                                    & PCB3        & 14.4  & 91.6                        & 33.7        & 96.8        & 51.1  & 98.1                        & 59.0       & 99.3        \\
    \multicolumn{1}{c|}{}                                    & PCB4        & 57.6  & 99.2                        & 66.5        & 99.4        & 71.4  & 98.7                        & 77.4       & 99.2        \\ \hline
    \multicolumn{1}{c|}{\multirow{4}{*}{Multiple instances}} & Macaroni1   & 32.9  & 99.8                        & 35.1        & 99.6        & 48.1  & 99.5                        & 52.6       & 99.9        \\
    \multicolumn{1}{c|}{}                                    & Macaroni2   & 16.0  & 95.8                        & 22.5        & 96.2        & 25.0  & 89.6                        & 30.2       & 94.9        \\
    \multicolumn{1}{c|}{}                                    & Capsules    & 74.2  & 98.1                        & 80.1        & 98.6        & 81.9  & 98.4                        & 90.2       & 99.2        \\
    \multicolumn{1}{c|}{}                                    & Candles     & 18.6  & 93.6                        & 22.7        & 93.4        & 46.5  & 98.1                        & 51.5       & 98.3        \\ \hline
    \multicolumn{1}{c|}{\multirow{4}{*}{Single instance}}    & Cashew      & 76.4  & 98.0                        & 83.3        & 99.5        & 85.5  & 97.5                        & 86.8       & 98.8        \\
    \multicolumn{1}{c|}{}                                    & Chewing gum & 84.1  & 99.4                        & 86.2        & 99.4        & 89.2  & 99.4                        & 90.1       & 99.3        \\
    \multicolumn{1}{c|}{}                                    & Fryum       & 81.9  & 98.7                        & 89.0        & 99.8        & 85.1  & 98.8                        & 85.4       & 98.6        \\
    \multicolumn{1}{c|}{}                                    & Pipe fryum  & 77.8  & 99.2                        & 81.1        & 99.6        & 86.1  & 97.8                        & 81.7       & 97.8        \\ \hline
    \multicolumn{1}{c|}{}                                    & Mean        &  53.8 & 97.3 & 59.7 & 98.1 & 65.1 & 97.3 & 71.9 & 98.5           \\ \hline
    \end{tabular}%
    }
\end{table*}

\begin{table*}[!ht]
	\centering
	\caption{2-class 5-shot anomaly detection on VisA.}
    \label{tab:2cls_cls_5shot_full}
    \vspace{-2mm}
	\setlength{\tabcolsep}{7pt}
	\resizebox{\linewidth}{!}{
	\centering
\begin{tabular}{cc|cccc|cccc}
\hline
                                                             &             & \multicolumn{2}{c|}{SimSiam}        & \multicolumn{2}{c|}{+SPD} & \multicolumn{2}{c|}{Supervised}     & \multicolumn{2}{c}{+SPD} \\ \hline
    \multicolumn{1}{c|}{}                                    &             & AU-PR & \multicolumn{1}{c|}{AU-ROC} & AU-PR       & AU-ROC      & AU-PR & \multicolumn{1}{c|}{AU-ROC} & AU-PR      & AU-ROC      \\ \hline
    \multicolumn{1}{c|}{\multirow{4}{*}{Complex structure}}  & PCB1        & 40.3  & 78.5                        & 47.0        & 89.8        & 55.6  & 91.7                        & 59.8       & 92.7        \\
    \multicolumn{1}{c|}{}                                    & PCB2        & 47.3  & 74.8                        & 46.4        & 79.2        & 42.4  & 77.6                        & 65.0       & 83.7        \\
    \multicolumn{1}{c|}{}                                    & PCB3        & 27.4  & 71.0                        & 24.7        & 68.0        & 24.5  & 67.4                        & 30.6       & 71.5        \\
    \multicolumn{1}{c|}{}                                    & PCB4        & 51.4  & 92.7                        & 71.3        & 96.6        & 77.9  & 96.6                        & 69.8       & 95.9        \\ \hline
    \multicolumn{1}{c|}{\multirow{4}{*}{Multiple instances}} & Macaroni1   & 43.8  & 86.9                        & 51.9        & 89.9        & 50.3  & 90.6                        & 40.8       & 85.9        \\
    \multicolumn{1}{c|}{}                                    & Macaroni2   & 12.5  & 59.3                        & 12.2        & 58.1        & 13.2  & 61.2                        & 14.8       & 63.4        \\
    \multicolumn{1}{c|}{}                                    & Capsules    & 33.6  & 69.8                        & 32.3        & 65.9        & 32.9  & 65.6                        & 29.3       & 63.2        \\
    \multicolumn{1}{c|}{}                                    & Candles     & 53.8  & 88.9                        & 67.5        & 92.0        & 67.7  & 92.5                        & 69.7       & 93.0        \\ \hline
    \multicolumn{1}{c|}{\multirow{4}{*}{Single instance}}    & Cashew      & 79.9  & 96.9                        & 80.2        & 96.8        & 81.0  & 96.8                        & 80.7       & 96.8        \\
    \multicolumn{1}{c|}{}                                    & Chewing gum & 49.1  & 72.5                        & 52.3        & 76.2        & 74.1  & 89.0                        & 67.5       & 87.5        \\
    \multicolumn{1}{c|}{}                                    & Fryum       & 97.4  & 99.2                        & 98.6        & 99.6        & 97.8  & 99.4                        & 96.8       & 99.3        \\
    \multicolumn{1}{c|}{}                                    & Pipe fryum  & 86.6  & 96.7                        & 88.5        & 96.3        & 93.5  & 98.1                        & 92.3       & 97.6        \\ \hline
    \multicolumn{1}{c|}{}                                    & Mean        &  51.9 & 82.3 & 56.1 & 84.0 & 59.2 & 85.5 & 59.8 & 85.9            \\ \hline
    \end{tabular}%
    }
\end{table*}

\begin{table*}[!ht]
	\centering
	\caption{2-class 10-shot anomaly detection on VisA.}
    \label{tab:2cls_cls_10shot_full}
    \vspace{-2mm}
	\setlength{\tabcolsep}{7pt}
	\resizebox{\linewidth}{!}{
	\centering
    \begin{tabular}{cc|cccc|cccc}
    \hline
                                                             &             & \multicolumn{2}{c|}{SimSiam}        & \multicolumn{2}{c|}{+SPD} & \multicolumn{2}{c|}{Supervised}     & \multicolumn{2}{c}{+SPD} \\ \hline
    \multicolumn{1}{c|}{}                                    &             & AU-PR & \multicolumn{1}{c|}{AU-ROC} & AU-PR       & AU-ROC      & AU-PR & \multicolumn{1}{c|}{AU-ROC} & AU-PR      & AU-ROC      \\ \hline
    \multicolumn{1}{c|}{\multirow{4}{*}{Complex structure}}  & PCB1        & 74.7  & 95.5                        & 74.5        & 95.5        & 68.0  & 94.2                        & 76.7       & 96.1        \\
    \multicolumn{1}{c|}{}                                    & PCB2        & 53.7  & 79.4                        & 60.7        & 85.8        & 73.3  & 89.2                        & 71.5       & 88.7        \\
    \multicolumn{1}{c|}{}                                    & PCB3        & 37.1  & 81.0                        & 48.2        & 87.9        & 47.4  & 86.2                        & 41.2       & 82.9        \\
    \multicolumn{1}{c|}{}                                    & PCB4        & 62.1  & 95.6                        & 73.0        & 97.0        & 76.2  & 97.0                        & 71.4       & 96.6        \\ \hline
    \multicolumn{1}{c|}{\multirow{4}{*}{Multiple instances}} & Macaroni1   & 59.2  & 92.3                        & 60.9        & 93.5        & 69.9  & 95.3                        & 68.4       & 94.2        \\
    \multicolumn{1}{c|}{}                                    & Macaroni2   & 19.9  & 65.6                        & 18.3        & 67.9        & 16.8  & 69.5                        & 30.4       & 76.9        \\
    \multicolumn{1}{c|}{}                                    & Capsules    & 50.2  & 82.0                        & 45.9        & 78.5        & 48.8  & 80.4                        & 45.1       & 80.6        \\
    \multicolumn{1}{c|}{}                                    & Candles     & 69.7  & 93.8                        & 72.5        & 94.0        & 79.2  & 95.7                        & 79.5       & 95.9        \\ \hline
    \multicolumn{1}{c|}{\multirow{4}{*}{Single instance}}    & Cashew      & 78.3  & 96.6                        & 79.0        & 96.8        & 78.5  & 96.6                        & 81.4       & 97.1        \\
    \multicolumn{1}{c|}{}                                    & Chewing gum & 80.4  & 92.7                        & 83.7        & 94.2        & 91.2  & 97.1                        & 92.9       & 97.3        \\
    \multicolumn{1}{c|}{}                                    & Fryum       & 98.4  & 99.3                        & 98.1        & 99.5        & 98.7  & 99.6                        & 98.5       & 99.7        \\
    \multicolumn{1}{c|}{}                                    & Pipe fryum  & 96.8  & 99.5                        & 96.5        & 99.4        & 96.9  & 99.3                        & 97.3       & 99.5        \\ \hline
    \multicolumn{1}{c|}{}                                    & Mean        &   65.0    &  89.4 & 67.6 & 90.8 & 70.4 & 91.7 & 71.2 & 92.1           \\ \hline
    \end{tabular}%
    }
\end{table*}

\begin{table*}[!ht]
	\centering
	\caption{2-class 5-shot anomaly segmentation on VisA.}
    \label{tab:2cls_seg_5shot_full}
    \vspace{-2mm}
	\setlength{\tabcolsep}{7pt}
	\resizebox{\linewidth}{!}{
	\centering
    \begin{tabular}{cc|cccc|cccc}
    \hline
                                                             &             & \multicolumn{2}{c|}{SimSiam}        & \multicolumn{2}{c|}{+SPD} & \multicolumn{2}{c|}{Supervised}     & \multicolumn{2}{c}{+SPD} \\ \hline
    \multicolumn{1}{c|}{}                                    &             & AU-PR & \multicolumn{1}{c|}{AU-ROC} & AU-PR       & AU-ROC      & AU-PR & \multicolumn{1}{c|}{AU-ROC} & AU-PR      & AU-ROC      \\ \hline
    \multicolumn{1}{c|}{\multirow{4}{*}{Complex structure}}  & PCB1        & 13.4  & 69.8                        & 13.8        & 73.3        & 2.3   & 66.8                        & 2.4        & 66.4        \\
    \multicolumn{1}{c|}{}                                    & PCB2        & 2.9   & 69.2                        & 1.0         & 67.6        & 2.5   & 55.3                        & 2.9        & 64.7        \\
    \multicolumn{1}{c|}{}                                    & PCB3        & 8.5   & 69.8                        & 11.3        & 65.0        & 14.4  & 72.7                        & 9.7        & 66.9        \\
    \multicolumn{1}{c|}{}                                    & PCB4        & 34.9  & 81.4                        & 31.9        & 82.3        & 31.1  & 83.1                        & 39.1       & 84.1        \\ \hline
    \multicolumn{1}{c|}{\multirow{4}{*}{Multiple instances}} & Macaroni1   & 2.3   & 80.1                        & 6.5         & 83.8        & 8.7   & 84.8                        & 7.1        & 86.4        \\
    \multicolumn{1}{c|}{}                                    & Macaroni2   & 0.2   & 80.4                        & 1.1         & 81.9        & 0.7   & 80.0                        & 0.3        & 80.2        \\
    \multicolumn{1}{c|}{}                                    & Capsules    & 11.3  & 68.5                        & 14.0        & 67.0        & 7.3   & 63.1                        & 12.4       & 70.1        \\
    \multicolumn{1}{c|}{}                                    & Candles     & 5.8   & 75.9                        & 7.2         & 71.0        & 4.3   & 73.8                        & 6.6        & 73.8        \\ \hline
    \multicolumn{1}{c|}{\multirow{4}{*}{Single instance}}    & Cashew      & 24.9  & 78.6                        & 23.3        & 87.2        & 21.6  & 77.3                        & 22.8       & 76.9        \\
    \multicolumn{1}{c|}{}                                    & Chewing gum & 70.0  & 90.7                        & 70.7        & 93.0        & 72.7  & 96.7                        & 71.0       & 96.0        \\
    \multicolumn{1}{c|}{}                                    & Fryum       & 6.3   & 67.2                        & 5.8         & 58.3        & 3.9   & 55.7                        & 7.5        & 62.1        \\
    \multicolumn{1}{c|}{}                                    & Pipe fryum  & 26.9  & 71.1                        & 31.6        & 81.3        & 44.4  & 86.2                        & 42.2       & 83.2        \\ \hline
    \multicolumn{1}{c|}{}                                    & Mean        & 17.3  & 75.2                        & 18.2        & 76.0        & 17.8  & 74.6                        & 18.7       & 75.9        \\ \hline
    \end{tabular}%
    }
\end{table*}

\begin{table*}[!ht]
	\centering
	\caption{2-class 10-shot anomaly segmentation on VisA.}
    \label{tab:2cls_seg_10shot_full}
    \vspace{-2mm}
	\setlength{\tabcolsep}{7pt}
	\resizebox{\linewidth}{!}{
	\centering
    \begin{tabular}{cc|cccc|cccc}
    \hline
                                                             &             & \multicolumn{2}{c|}{SimSiam}        & \multicolumn{2}{c|}{+SPD} & \multicolumn{2}{c|}{Supervised}     & \multicolumn{2}{c}{+SPD} \\ \hline
    \multicolumn{1}{c|}{}                                    &             & AU-PR & \multicolumn{1}{c|}{AU-ROC} & AU-PR       & AU-ROC      & AU-PR & \multicolumn{1}{c|}{AU-ROC} & AU-PR      & AU-ROC      \\ \hline
    \multicolumn{1}{c|}{\multirow{4}{*}{Complex structure}}  & PCB1        & 17.1  & 72.1                        & 24.5        & 80.7        & 6.5   & 66.5                        & 5.3        & 58.5        \\
    \multicolumn{1}{c|}{}                                    & PCB2        & 12.5  & 63.3                        & 7.7         & 81.7        & 11.4  & 72.9                        & 14.3       & 75.7        \\
    \multicolumn{1}{c|}{}                                    & PCB3        & 23.3  & 80.2                        & 18.1        & 73.6        & 21.7  & 75.8                        & 25.5       & 78.6        \\
    \multicolumn{1}{c|}{}                                    & PCB4        & 45.2  & 92.1                        & 46.9        & 86.2        & 41.3  & 88.6                        & 50.1       & 91.4        \\ \hline
    \multicolumn{1}{c|}{\multirow{4}{*}{Multiple instances}} & Macaroni1   & 10.3  & 83.2                        & 12.8        & 86.2        & 20.8  & 92.4                        & 14.4       & 88.1        \\
    \multicolumn{1}{c|}{}                                    & Macaroni2   & 7.3   & 89.0                        & 7.0         & 78.4        & 8.8   & 87.6                        & 8.8        & 87.4        \\
    \multicolumn{1}{c|}{}                                    & Capsules    & 36.1  & 88.0                        & 42.4        & 83.7        & 36.7  & 83.6                        & 29.9       & 74.9        \\
    \multicolumn{1}{c|}{}                                    & Candles     & 11.6  & 71.9                        & 18.0        & 82.0        & 13.1  & 79.9                        & 14.9       & 84.0        \\ \hline
    \multicolumn{1}{c|}{\multirow{4}{*}{Single instance}}    & Cashew      & 32.0  & 85.1                        & 30.1        & 86.6        & 43.7  & 84.4                        & 37.1       & 86.2        \\
    \multicolumn{1}{c|}{}                                    & Chewing gum & 76.0  & 96.0                        & 78.9        & 94.8        & 77.0  & 97.5                        & 81.7       & 97.1        \\
    \multicolumn{1}{c|}{}                                    & Fryum       & 29.7  & 74.3                        & 27.9        & 76.5        & 20.2  & 70.5                        & 32.2       & 69.3        \\
    \multicolumn{1}{c|}{}                                    & Pipe fryum  & 40.5  & 84.5                        & 42.4        & 88.5        & 38.9  & 81.3                        & 53.3       & 90.9        \\ \hline
    \multicolumn{1}{c|}{}                                    & Mean        & 28.5  & 81.6                        & 29.7        & 83.2        & 28.3  & 81.8                        & 30.6       & 81.8        \\ \hline
    \end{tabular}%
    }
\end{table*}

\section{Qualitative Results}

\begin{figure}[!t]
\centering
\includegraphics[width=\textwidth]{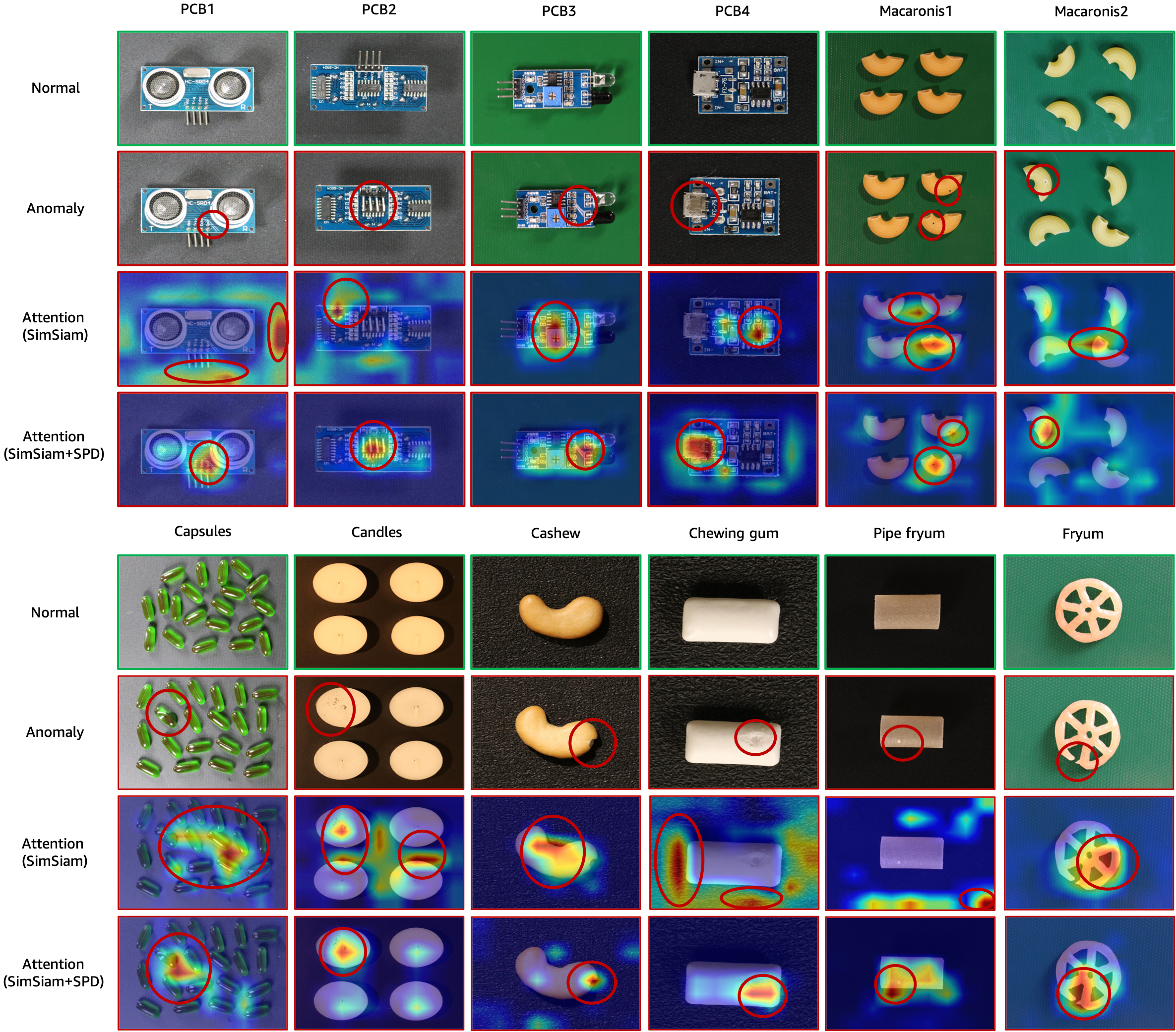}
 \caption{Attention maps generated by GradCAM. 1st row: normal images; 2nd row: anomalous images; 3rd row: GradCAM based on SimSiam; 4th row: GradCAM based on SimSiam+SPD. Defects and high energy (red) parts in attentions are highlighted. Best viewed by zooming in.}
 \label{fig:diff_attention}
\end{figure}

\noindent\textbf{Attention maps: }
In Fig. \ref{fig:diff_attention}, we show the qualitative results to demonstrate the effectiveness of SPD regularization. Based on GradCAM \cite{selvaraju2017grad}, we generate attention maps of anomalous images by regarding negative cosine similarity as the distance (loss) between the defective image and its nearest normal sample. High energy regions contribute mostly to the feature cosine distance. Compared to SimSiam, SPD helps the baseline model to be more sensitive to the defective regions, demonstrating the validity of proposed SPD learning.

\begin{figure}[!t]
  \centering
  \includegraphics[width=\linewidth]{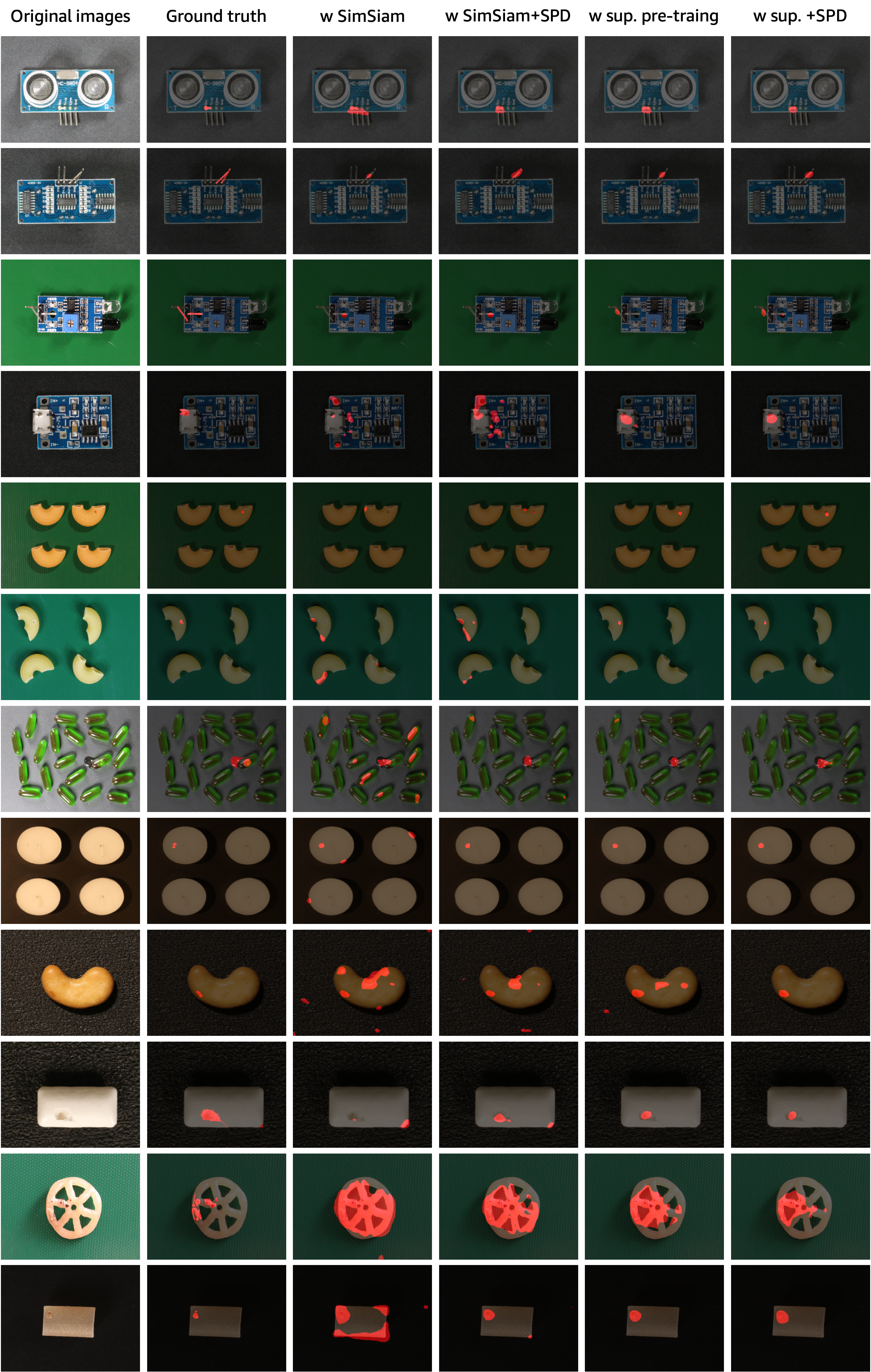}
  \vspace{-.75cm}
  \caption{Segmentation results from PaDiM with various pre-trained models.}
  \label{fig:1cls_seg}
\end{figure}

\noindent\textbf{Anomaly segmentation results: }
In Fig. \ref{fig:1cls_seg}, we show the segmentation results for PaDiM with SimSiam, SimSiam+SPD, supervised and supervised+SPD pre-trained ResNet-50. We can see the SPD gives better qualitative segmentation results than each baseline.

\clearpage

\end{document}